\documentclass{article}

\usepackage{microtype}
\usepackage{graphicx}
\usepackage{subcaption}
\usepackage{booktabs} 
\usepackage{bbding}
\usepackage{hyperref}
\usepackage{soul}
\usepackage{tcolorbox}
\usepackage{fontawesome}  
\usepackage{listings}
\tcbuselibrary{breakable}

\definecolor{lightpink}{RGB}{255, 230, 240}
\definecolor{lightblue}{RGB}{230, 240, 255}
\definecolor{lightgray}{RGB}{248, 248, 248}
\definecolor{bordergray}{RGB}{200, 200, 200}

\definecolor{bestcolor}{RGB}{219, 208, 237}
\definecolor{secondcolor}{RGB}{241, 237, 248}
\definecolor{thirdcolor}{RGB}{211, 222, 190}
\definecolor{line-blue}{RGB}{243, 248, 252}
\definecolor{ForestGreen}{RGB}{34, 139, 34}

\usepackage[accepted]{icml2026}
\usepackage{amsmath}
\usepackage{amssymb}
\usepackage{mathtools}
\usepackage{amsthm}
\usepackage[table]{xcolor}

\usepackage[capitalize,noabbrev]{cleveref}

\theoremstyle{plain}

\theoremstyle{definition}

\theoremstyle{remark}

\definecolor{bestcol}{HTML}{A8D8B9}
\definecolor{secondcol}{HTML}{D9F2E1}

\icmltitlerunning{PerceptionRubrics: Calibrating Multimodal Evaluation to Human Perception}

\begin{document}

\twocolumn[
  \icmltitle{PerceptionRubrics: Calibrating Multimodal Evaluation to Human Perception}



  \icmlsetsymbol{core}{*}

  \begin{icmlauthorlist}
    \icmlauthor{Yana Wei}{core,jhu}
    \icmlauthor{Hongbo Peng}{core,stepfun}
    \icmlauthor{Yanlin Lai}{core,tsinghua}
    \icmlauthor{Liang Zhao}{stepfun}
    \icmlauthor{Kangheng Lin}{stepfun}
    \icmlauthor{En Yu}{stepfun}
    \icmlauthor{Keyu Lv}{stepfun}
    \icmlauthor{Han Zhou}{stepfun}
    \icmlauthor{Yin Tang}{independent}
    \icmlauthor{Haodong Li}{stepfun}
    \icmlauthor{Mitt Huang}{stepfun}
    \icmlauthor{Hangyu Guo}{stepfun}
    \icmlauthor{Jianjian Sun}{stepfun}
    \icmlauthor{Zheng Ge}{stepfun}
    \icmlauthor{Xiangyu Zhang}{stepfun}
    \icmlauthor{Daxin Jiang}{stepfun}
    \icmlauthor{Vishal M. Patel}{jhu}
  \end{icmlauthorlist}

  \icmlaffiliation{jhu}{Johns Hopkins University}
  \icmlaffiliation{stepfun}{StepFun}
  \icmlaffiliation{tsinghua}{Tsinghua University}
  \icmlaffiliation{independent}{Independent Researcher}

\icmlcorrespondingauthor{Yana Wei}{ywei66@jh.edu}
\icmlcorrespondingauthor{Vishal M. Patel}{vpatel36@jhu.edu}



  \icmlkeywords{Machine Learning, ICML}

  \vskip 0.3in
]



\printAffiliationsAndNotice{\textsuperscript{*}Core contribution.}

\begin{abstract}

We introduce \textsc{PerceptionRubrics}, a rubric-based evaluation framework that addresses the gap between saturated benchmark scores and real-world brittleness. Shifting evaluation from holistic semantic matching to rigorous atomic auditing, \textsc{PerceptionRubrics} pairs 1,038 information-dense images with over 10,000 instance-specific rubrics. These criteria are derived from golden captions constructed via a novel Circular Peer-Review consensus pipeline and then distilled into a dual-stream system of \emph{Must-Right} (essential facts) and \emph{Easy-Wrong} (fine-grained details) rubrics. Crucially, \textsc{PerceptionRubrics} implements a Gated Scoring mechanism: unlike linear averages, failure on mandatory visual facts triggers sharp binary penalties. Extensive evaluation yields critical insights: (1) \textbf{The Reliability Gap}: models often verify fragmented elements correctly yet fail strict conjunctive constraints, exposing brittleness in dense domains; (2) \textbf{Open-Closed Stratification}: contrary to reasoning trends, we reveal a persistent 8\% perception deficit between open-source and proprietary frontiers; and (3) \textbf{Human-Aligned Rigor}: our gated metrics substantially out-align conventional benchmarks, validating that strict perceptual fidelity is the prerequisite for reliable generation. Code and data
can be found at \href{https://weiyana.github.io/PerceptionRubrics}{\texttt{project page}}.

\end{abstract}
\section{Introduction}
\begin{figure}[h]
\includegraphics[width=1.0\columnwidth]{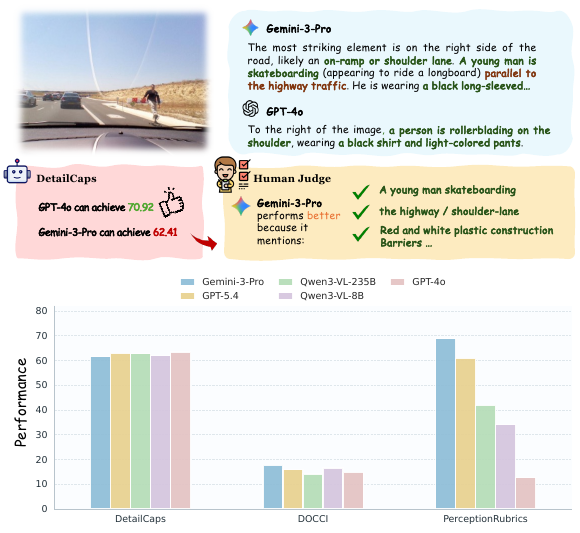}
\caption{
\textbf{Motivation of \textsc{PerceptionRubrics}. }
\textbf{Top:} An existing benchmark favors GPT-4o despite key omissions, while humans prefer responses that capture more perceptually important details. 
\textbf{Bottom:} Compared with DetailCaps and DOCCI, \textsc{PerceptionRubrics} more clearly distinguishes model capabilities.
}
\label{fig:adv}
\vspace{-4mm}
\end{figure}


Despite the rapid evolution of Multimodal Large Language Models (MLLMs), a fundamental evaluation crisis persists: \textbf{current perception benchmarks do not reliably reflect genuine perceptual capability}. This has led to a evaluation paradox where leaderboards are increasingly saturated in the high-score regime as illustrated in~\cref{fig:adv}, yet models remain perceptually brittle in real-world deployment. Top-tier systems often appear nearly tied on metrics but exhibit drastically different failure modes---such as miscounting objects or inverting spatial relations---that are highly salient to users even when reported metric scores \citep{capture} remain high. This discrepancy suggests that benchmark rewards are \textbf{misaligned with human perceptual sensitivity}, creating a false sense of progress and failing to provide the diagnostic resolution needed to steer the next generation of MLLMs.

\begin{figure*}[t]
    \centering
    \includegraphics[width=0.99\linewidth]{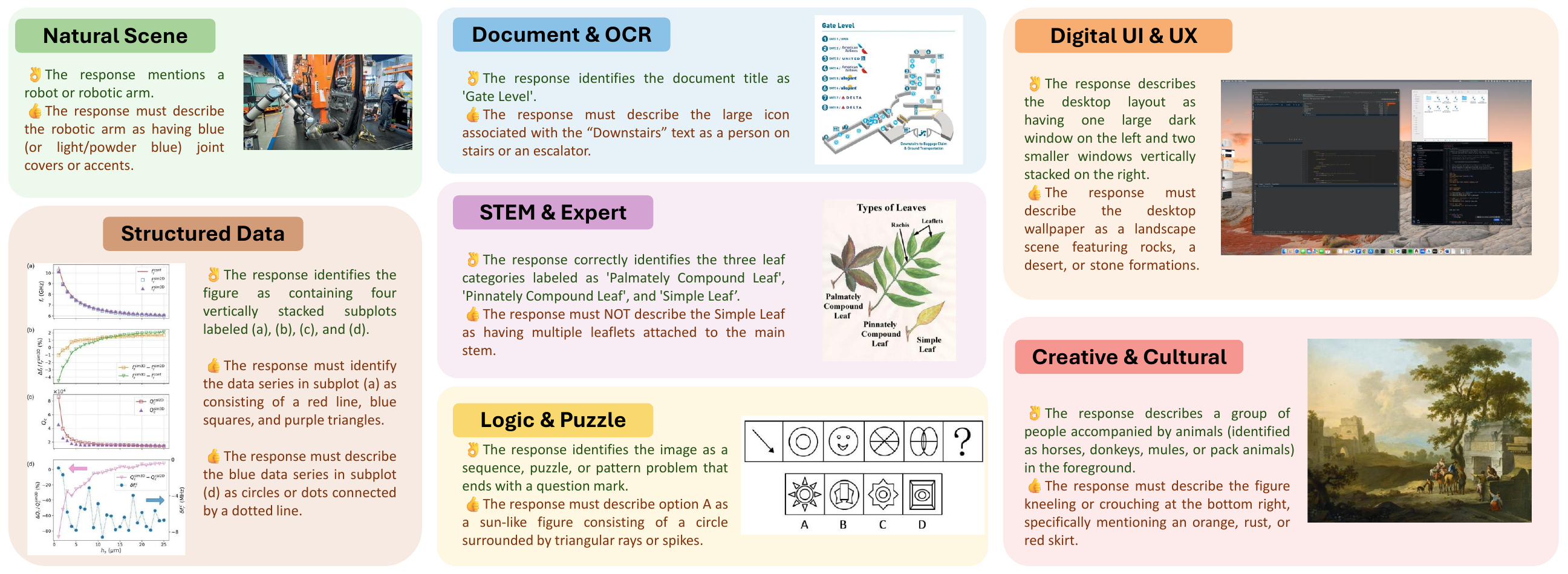}
    \caption{\textbf{Rubric Demonstration of \textsc{PerceptionRubrics}.} Representative examples are selected for each task, highlighting ``Must Right'' (\includegraphics[height=1em]{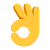}; essential features) and ``Easy Wrong'' pitfalls (\includegraphics[height=1em]{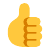}; error-prone fine-grained details).
    }
    \label{fig:main_data}
\end{figure*}

We trace this failure to two systemic flaws in current benchmark design. \textbf{First, the visual content and task design lack sufficient perceptual detail coverage.} Many benchmarks rely on information-poor images or narrow domains~\citep{onoe2024docci}, often framing tasks as closed-form questions that allow models to ``shortcut'' through linguistic priors rather than genuine visual grounding~\citep{zhou2023analyzing,zhang2025mitigating}. Even in open-ended captioning, references are frequently imprecise, biased, or too sparse \citep{capture} to challenge the long-tail visual knowledge of frontier models. \textbf{Second, current reward signals are fundamentally uncalibrated.} Conventional metrics, such as single-number similarity scores (e.g., CLIPScore~\citep{radford2021clip}) or averaged multi-aspect schemes~\citep{capture}, rely on linear averaging that effectively ``dilutes'' fatal localized errors with general semantic overlap. Consequently, a caption plagued by hallucinations can still achieve a high metric score, severing the link between numerical performance and genuine reliability. In contrast, human perception is strictly \textbf{non-linear}: a single-digit hallucination in a financial table is not a permissible fluctuation but a binary failure~\citep{poznanski2025olmocr}. Existing metrics fail to reflect this, making it difficult to distinguish acceptable descriptive variation from critical perceptual failures.


To bridge this gap, we propose \textsc{PerceptionRubrics}, a benchmark that repurposes image captioning---the most fundamental proxy for integrated perception, recognition, and reasoning---into a rigorous diagnostic testbed. To address the \textit{data deficit}, we curate 1,038 images characterized by extreme information density and distributional diversity. Crucially, to bypass the visual grounding gap that limits direct image-to-rubric generation, we adopt a \textbf{caption-centric construction pipeline} as an intermediary strategy. Instead of relying on noisy raw predictions, we establish ground truth via a \textbf{Circular Peer-Review} consensus mechanism: an ensemble of state-of-the-art MLLMs iteratively critiques and refines descriptions, followed by human verification. This process yields ``Golden Captions'' that serve as high-fidelity textual references for the visual content, filtering out the noise and biases prevalent in traditional datasets.

Building on this foundation, we address the \textit{calibration gap} by distilling Golden Captions into a rubric-based auditing system. With human refinement, we extract over 10,000 atomic rubrics and organize them into two complementary streams: \emph{Must-Right} rubrics, which capture essential visual facts that a response must satisfy, and \emph{Easy-Wrong} rubrics, which target common hallucinations, omissions, and misinterpretations mined from model error patterns. 
We then introduce a gated scoring mechanism calibrated to human sensitivity: the Must-Right rubrics serve as mandatory gatekeepers, so failure to satisfy any essential criterion sharply penalizes the final score. 
This design ensures that the metric reflects not just coarse semantic proximity, but genuine perceptual reliability, effectively distinguishing between acceptable approximations and catastrophic failures.

Comprehensive evaluation and analysis across leading MLLMs on \textsc{PerceptionRubrics} yields critical insights:

\begin{itemize}
    \item \textbf{Unveiling the ``Reliability Gap''.} We expose a disconnect between \textit{fragmented recognition} and \textit{coherent understanding}: models often pass atomic checks but fail strict conjunctive constraints. This reveals that despite high partial scores, current MLLMs lack the perceptual consistency required for information-dense domains like GUIs.

    \item \textbf{Quantifying the Open-Closed Gap.} Contrasting the convergence in reasoning tasks, we identify a persistent 8\% perception deficit between the open-source frontier (e.g., Qwen3.5~\citep{qwen35blog}) and proprietary leaders (e.g., Seed-2.0~\citep{Seed2}). Basic visual precision thus remains a decisive bottleneck distinguishing intrinsic model capacity.

    \item \textbf{Superior Human Alignment.} \textsc{PerceptionRubrics} aligns substantially better with human judgment than conventional benchmarks (e.g., DOCCI~\citep{onoe2024docci}), an effect amplified by our gated scoring. Furthermore, a near-perfect correlation between basic perception and hallucination resistance confirms strict fidelity as a prerequisite for reliable generation.
\end{itemize}

\section{Related Work}

\paragraph{Visual Perception Benchmarks in MLLMs.} Evaluating visual perception remains pivotal for assessing MLLMs~\citep{gemini3pro,gpt5p2}. Current benchmarks generally fall into two categories: holistic suites and task-specific datasets. Comprehensive frameworks like MMBench~\citep{mmbench}, MM-Vet~\citep{mmvet}, and MME~\citep{mme} evaluate broad capabilities but increasingly face leaderboard saturation in recent flagship models~\citep{bai2025qwen3vltechnicalreport,step3vl10b}. Conversely, task-specific benchmarks target distinct skills, such as OCR in OCRBench~\citep{ocrbench}, open-world recognition in SimpleVQA~\citep{cheng2025simplevqamultimodalfactualityevaluation} and spatial understanding in VSIBench~\citep{vsibench}. However, these benchmarks heavily rely on closed-ended formats (e.g., single or multiple-choice). Such designs often allow models to exploit linguistic priors or random guessing to bypass genuine visual grounding~\citep{zhou2023analyzing,zhang2025mitigating}, limiting their ability to diagnose perceptual brittleness.

\begin{figure}[t]
    \centering
    \includegraphics[width=1.0\linewidth]{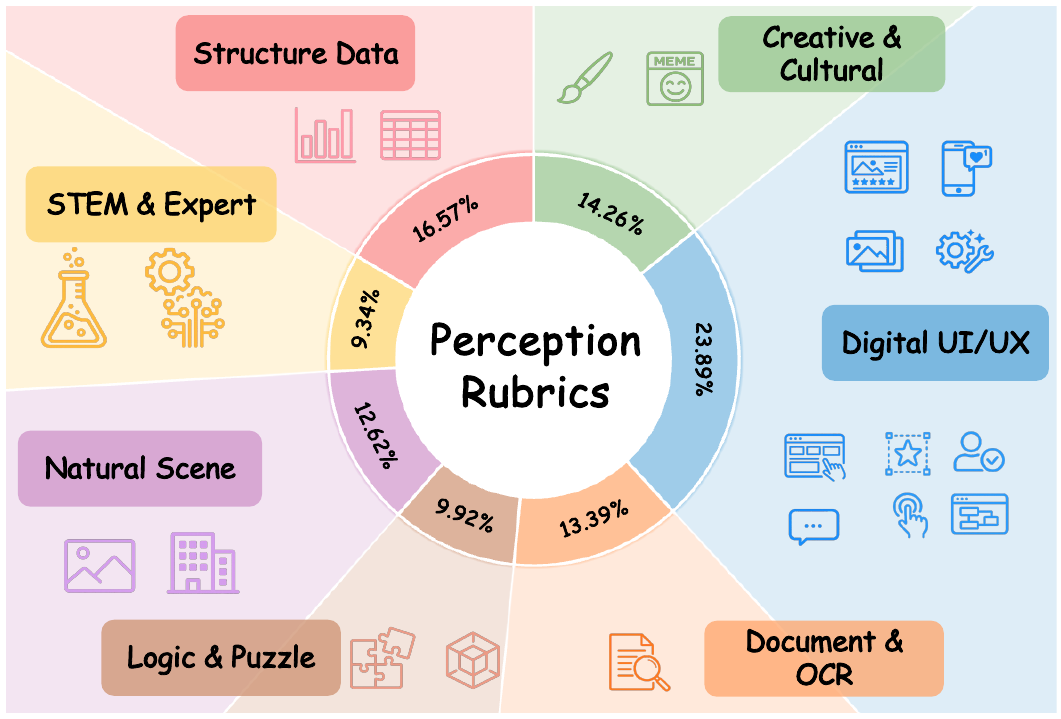}
    \caption{\textbf{Benchmark Statistics of \textsc{PerceptionRubrics}:} The distribution of tasks across 7 main categories.}
    \label{fig:data}
\end{figure}

\begin{figure*}[t]
    \centering
    \includegraphics[width=0.95\linewidth]{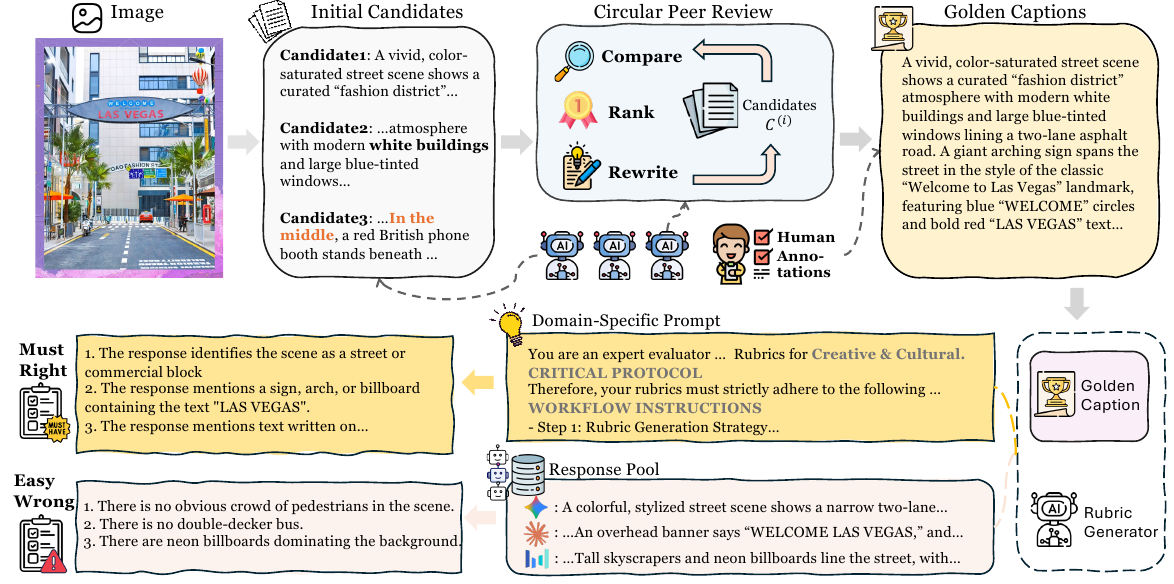}
    \caption{\textbf{The \textsc{PerceptionRubrics} Construction Pipeline.} Adopting a caption-centric approach, we first synthesize golden captions via circular peer-review (Top). These captions then serve as anchors to generate Must-Right and Easy-Wrong rubrics through domain-specific prompting (Bottom).}
    \label{fig:caption}
\vspace{-3mm}

\end{figure*}

\paragraph{Evaluation of Image Captioning.} Image captioning serves as a holistic proxy for perception, requiring models to autonomously prioritize and describe visual elements. Recent methods have moved beyond generic similarity metrics~\citep{papineni2002bleu} or object-set matching heuristics~\citep{rohrbach2018object} towards model-based evaluation. DOCCI~\citep{onoe2024docci} targets detailed description using reference-based metrics; DetailCaps~\citep{capture} employs multi-expert annotation to score object and attribute matching; RePer~\citep{wei2025perception} utilizes an LLM-judge for aspect-based evaluation; and CapArena~\citep{cheng2025caparena} aligns assessments with human preference via pairwise battles. Despite these advancements, a critical gap persists: existing methods often rely on sparse, biased references and linear scoring mechanisms that dilute fatal localized hallucinations with high holistic similarity, failing to reflect the non-linear sensitivity of human verification~\citep{poznanski2025olmocr}.

\paragraph{Rubric-Based Reward Modeling.} To improve evaluation reliability, the field is shifting from opaque scalar scoring~\citep{liu2024skywork} to rubric-based auditing. In text generation, structured criteria have effectively mitigated reward hacking~\citep{rezaei2025online}. Approaches like RM-R1~\citep{chen2025rm} and SPCT~\citep{liu2025inference} formulate evaluation as a reasoning process via chain-of-rubrics, while frameworks such as RaR~\citep{gunjal2025rubrics} and ResearchRubrics\citep{sharma2025researchrubrics} leverage LLMs to decompose subjective judgments into atomic, verifiable checks. While this paradigm has standardized text-centric evaluation, comparable fine-grained auditing systems for multimodal perception remain under-explored. Existing vision benchmarks lack the mechanism to decompose complex visual scenes into verifiable atomic facts, highlighting the need for a rigorous standard to distinguish precise perception from approximation.
\section{PerceptionRubrics}

To align multimodal evaluation with the rigor of human judgment, we first outline our guiding design principles (\cref{sec:design}) and data curation strategy (\cref{sec:curation}), followed by our novel caption-centric pipeline for generating atomic rubrics (\cref{sec:rubric_construction}) and the gated scoring mechanism that enforces calibration (\cref{sec:metric}).

\subsection{Design Criteria}\label{sec:design}
To rigorously stress-test the upper bounds of state-of-the-art models and bridge the gap between reported metrics and real-world reliability, the design of \textsc{PerceptionRubrics} is governed by two overarching principles:

\paragraph{Enforcing Perceptual Persistence.} To probe comprehensive perceptual capabilities, we prioritize complexity over scale. We posit that a robust benchmark must utilize images with extreme information density that ranging from crowded scenes to document-heavy layouts, therefore invalidate the linguistic ``shortcuts'' often taken by models. This design criterion compels models to exhibit \emph{perceptual persistence}, requiring active, fine-grained exploration of long-tail visual details rather than reliance on rough global understanding or parametric priors.

\paragraph{Calibrating to Human Sensitivity.} To resolve the paradox where high semantic scores mask brittle performance, we prioritize precision over approximation. We argue that an effective metric must mirror the \emph{error-sensitive} nature of human judgment, where localized errors (e.g., hallucinating a single digit in a chart) represent binary failures rather than minor fluctuations. Consequently, our criterion mandates atomic verifiability and task-adaptive penalties: evaluation must be grounded in objective, fact-based checks (True/False) and rigorously penalize hallucinations, ensuring the metric reflects practical perceptual utility rather than mere statistical similarity.

\subsection{Image Curation}\label{sec:curation}

To ensure the benchmark probes the perceptual limits of flagship models, we curate an image collection that emphasizes visual diversity and complexity, targeting inputs rich in perceptually critical details that maximize error potential.

\paragraph{Task Domains.} As illustrated in \cref{fig:data}, we structure our data across seven diverse categories to cover the full spectrum of multimodal capabilities: \emph{Natural Scenes} (complex real-world environments); \emph{Document \& OCR} (text-dense documents, forms, and handwritten content); \emph{Digital UI \& UX} (web pages, mobile UIs, and dashboards); \emph{Structured Data} (charts, plots, and tables); \emph{STEM \& Expert} (scientific diagrams, geometric figures, and medical imaging); \emph{Logic \& Puzzle} (visual riddles and spatial reasoning tasks); and \emph{Creative \& Cultural} (artworks, cultural artifacts, and design concepts).

\paragraph{Density-Aware Filtering.} We employ the advanced MLLM, Step3-VL-10B~\citep{step3vl10b}, as a scorer to filter the curated images based on complexity and informativeness. Specifically, given a candidate image, the model evaluates its visual complexity (via object richness) and informativeness (via semantic density), assigning a score from 1 to 10 (see details in \cref{prompt:complexity}). To ensure a balanced distribution across categories, we retain images that surpass domain-specific thresholds.

\subsection{Caption-Centric Perception Rubric Construction}
\label{sec:rubric_construction}

To instantiate the rigorous design criteria outlined above, we construct a caption-centric pipeline. 
Given that generating rubrics directly from raw pixels often suffers from the visual grounding gap inherent in current vision encoders~\citep{register} and MLLMs~\citep{visualattentionsink}, we choose an intermediary strategy: first explicitly transcribing visual information into text, then distilling rules from it.
This approach prioritizes constructing a comprehensive, precise, and exhaustive golden caption to capture image details. This textual foundation enables the subsequent rubric generator to cover extreme visual granularity and detect subtle failure modes with significantly higher reliability than direct image-to-rubric methods.

\subsubsection{Generating Golden Caption}
As illustrated in the top half of \cref{fig:caption}, we construct golden reference captions $C_{gold}$ through a two-step consensus-driven pipeline. This approach treats heterogeneous MLLMs as a collaborative filter to minimize human annotation costs while ensuring high precision.

\paragraph{Step 1: Circular Peer-Review.} 
Three distinct top-tier MLLMs (e.g., GPT-5.2, Gemini-3-Pro, and Seed-1.8) serve as a ``jury-and-generator'' ensemble. For each image, they first generate independent descriptions to form an initial candidate pool. To reduce hallucinations and self-preference bias, we implement a \textbf{circular peer-review mechanism} (\cref{fig:caption}, top middle). In this phase, models iteratively \textbf{compare} candidates against visual evidence, \textbf{rank} them based on accuracy, and \textbf{rewrite} descriptions to synthesize a superior version. This review cycle runs for limited iterations ($N \le 2$) to efficiently drive the ensemble toward a unified consensus.

\paragraph{Step 2: Strict Consensus Filtering.} 
To strictly control quality and annotation costs, human experts intervene only as final verifiers rather than creators. We adopt a \textbf{discard-on-divergence} protocol: samples where the models fail to reach a unanimous agreement are discarded. Only when the ensemble converges on a single optimal caption (i.e., high consensus) do human annotators perform a lightweight verification to finalize the golden reference $C_{gold}$. This ensures that human effort is spent exclusively on high-confidence samples.

\subsubsection{Generating Perception Rubric}
Building upon the verified golden reference $C_{gold}$, we employ Gemini-3-Pro~\citep{gemini3pro} as the rubric proposer to construct dual-stream evaluation criteria (\cref{fig:caption}, bottom). This pipeline mirrors the error-sensitive nature of human judgment by generating rubrics from two complementary perspectives: \textit{a priori} essential facts and \textit{a posteriori} common pitfalls.

\textbf{A Priori: \emph{Must-Right} Rubrics.} 
From a positive perspective, the rubric proposer distills a set of atomic perceptual facts from $I$ and $C_{\text{gold}}$ that a candidate \emph{must} correctly identify. Crucially, we employ \textbf{domain-specific adaptive prompts} (detailed in \cref{prompt:generate}) to align with varying perceptual demands: rubrics for text-centric images prioritize character precision, while those for natural scenes emphasize spatial relations and object attributes.

\textbf{A Posteriori: \emph{Easy-Wrong} Rubrics.} 
From a negative perspective, we challenge model robustness by targeting likely failure modes. We first construct a \textit{response pool} $\mathcal{P}$ by collecting predictions from a diverse set of baseline MLLMs. By analyzing the discrepancies between these actual outputs $\mathcal{P}$ and the reference $C_{gold}$, the rubric proposer identifies frequent hallucinations and subtle misinterpretations. These empirically observed errors are converted into Easy-Wrong rubrics, ensuring the evaluation penalizes realistic mistakes rather than hypothetical ones.

\subsection{Evaluation Metric}\label{sec:metric}

We employ an LLM-as-a-Judge framework to perform fine-grained evaluation, aiming to balance effectiveness and efficiency. We select GPT-OSS-120B~\citep{openai2025gptoss} as the judge due to its proven capability for highly calibrated assessments~\citep{step3vl10b}. Specifically, a model prediction $P$, and a set of rubrics $\mathcal{R} = \mathcal{R}_{m} \cup \mathcal{R}_{e}$ covering \emph{Must-Right} and \emph{Easy-Wrong} cases, the judge evaluates each rubric item yielding a boolean output (\textit{True} for compliance, \textit{False} otherwise). To prioritize factual correctness, we implement a gated scoring logic:

\paragraph{Must-Right as the Gate.} Let $\mathcal{R}_{m} = \{r_{m,1}, \dots, r_{m,j}\}$ be the set of Must-Right rubrics, which serve as a mandatory gatekeeper. If the model fails even a single criterion in $\mathcal{R}_{m}$, the description is deemed factually compromised, penalizing the final score to zero:
\begin{equation}
G = \prod_{i=1}^{j} \mathbb{I}(r_{m,i} = \text{True})
\end{equation}
where $G \in \{0, 1\}$ represents the gate status.

\paragraph{Easy-Wrong for Granular Differentiation.}
For models that pass the gate ($G=1$), we calculate the final score based on the Easy-Wrong rubrics $\mathcal{R}_{e} = \{r_{e,1}, \dots, r_{e,k}\}$. 
These rubrics assess whether the response correctly captures error-prone fine-grained details, including details that are commonly hallucinated, omitted, or misinterpreted.
The final score $S$ is defined as:
\begin{equation}
S = G \cdot \frac{1}{k} \sum_{i=1}^{k} \mathbb{I}(r_{e,i} = \text{True})
\end{equation}
This scoring philosophy ensures that a high score reflects not only the absence of basic hallucinations but also a superior discernment of subtle, density-rich visual details.






\begin{table}[t]
\centering
\small
\caption{Detailed statistics of the \textsc{PerceptionRubrics} benchmark for images, captions and rubrics.}
\label{tab:prb_stats}
\begin{tabular}{l c}
\toprule
\textbf{Statistic} & \textbf{Value} \\
\midrule
Number of images & 1{,}038 \\
Number of captions & 1{,}038 \\
Average caption length (words) & 770.42 \\ 
\midrule
Total number of rubrics & 10{,}718 \\
\quad Must-right rubrics & 4{,}053 \\ 
\quad Easy-wrong rubrics & 6{,}665 \\ 
\midrule
Average rubrics per image & 10.33 \\ 
\quad Must-right per image & 3.90 \\ 
\quad Easy-wrong per image & 6.42 \\ 
\bottomrule
\end{tabular}
\end{table}

\section{Experiments}


\subsection{Benchmark Statistics}
As summarized in \cref{tab:prb_stats},
the resulting benchmark contains 1,038 information-dense images, each paired with a verified golden caption and a set of instance-specific perception rubrics. In total, \textsc{PerceptionRubrics} includes 10,718 atomic rubrics, consisting of 4,053 Must-Right rubrics and 6,665 Easy-Wrong rubrics, with an average of 10.33 rubrics per image. Beyond rubric density, our benchmark is also characterized by highly detailed textual references. 
As shown in \cref{fig:word_count_dist}, the golden caption lengths exhibit a right-skewed distribution: most captions concentrate around 400--500 words, while a long tail extends to captions exceeding 3,400 words. The mean caption length reaches 770.42 words, higher than the median of 569 words. This long-tailed caption distribution reflects the high information density of our images and provides a rich textual anchor for constructing fine-grained and verifiable rubrics.

\subsection{Experimental Setup}
We evaluate a diverse suite of 25 models, spanning proprietary frontier models (e.g., Gemini-3-Pro~\citep{gemini3pro}, Gemini-3.5-Flash~\citep{gemini3.5flash}, GPT-5.4~\citep{gpt5.4}, GPT-4o~\citep{gpt4o}, Seed-2.0~\citep{Seed2}, Seed-1.8~\citep{Seed1p8}, Seed-1.6~\citep{Seed1p6}, GLM-5V-Turbo~\citep{glm5vturbo}, Qwen3.5-Plus~\citep{qwen35blog}) and leading open-weights models (e.g.,Qwen3.5-397B~\citep{qwen35blog}, Qwen3-VL~\citep{bai2025qwen3vltechnicalreport}, Qwen2.5-VL~\citep{qwen25vl},Step3-VL-10B~\citep{step3vl10b}, Step-3.7-Flash~\citep{step37flash}, MiniMax-M3~\citep{lai2026minimaxsparseattention}, MiMo-V2.5~\citep{mimov25},Kimi-K2.5~\citep{kimik2p5}). 

\subsection{Main Results}


\noindent\textbf{Compliance Scores.}
\cref{tab:main_results} summarizes the performance of all evaluated models, which reveals a pronounced performance stratification that is largely obscured by traditional holistic benchmarks. \textbf{Seed-2.0-Lite} leads the leaderboard with an overall score of $70.07\%$, outperforming the runner-up (\textbf{Gemini-3.5-Flash}) by $0.19\%$. 
In contrast, despite being a widely used proprietary model, \textbf{GPT-4o-2024-05-13} exhibits the weakest perceptual performance among its category, achieving an overall accuracy of only $12.59\%$. 
Across models, performance is consistently higher on \textbf{natural} image domains (e.g., reaching $79.20\%$ for Seed-2.0-Lite), aligning with human perceptual intuition and reflecting the relative maturity of models in handling real-world visual scenes. Conversely, almost all models struggle most in the \textbf{GUI} domain (e.g., Qwen2.5-VL-7B drops to $5.13\%$), indicating that \textit{robust visual grounding for future agents remains an unresolved challenge}. 
Moreover, unlike in reasoning tasks where open-sourced models often rival proprietary flagships~\citep{step3vl10b,bai2025qwen3vltechnicalreport}, our results show a distinctive performance gap. The best-performing open-source model (\textbf{Qwen3.5}, $61.61\%$) still trails the proprietary state-of-the-art by over $8\%$. This suggests that \textit{open-source models still have significant ground to cover in fine-grained perception and open-world recognition}, also confirming our benchmark's sensitivity in distinguishing intrinsic model capacity beyond reasoning capabilities.



\begin{figure}[t]
    \centering
    \includegraphics[width=0.98\linewidth]{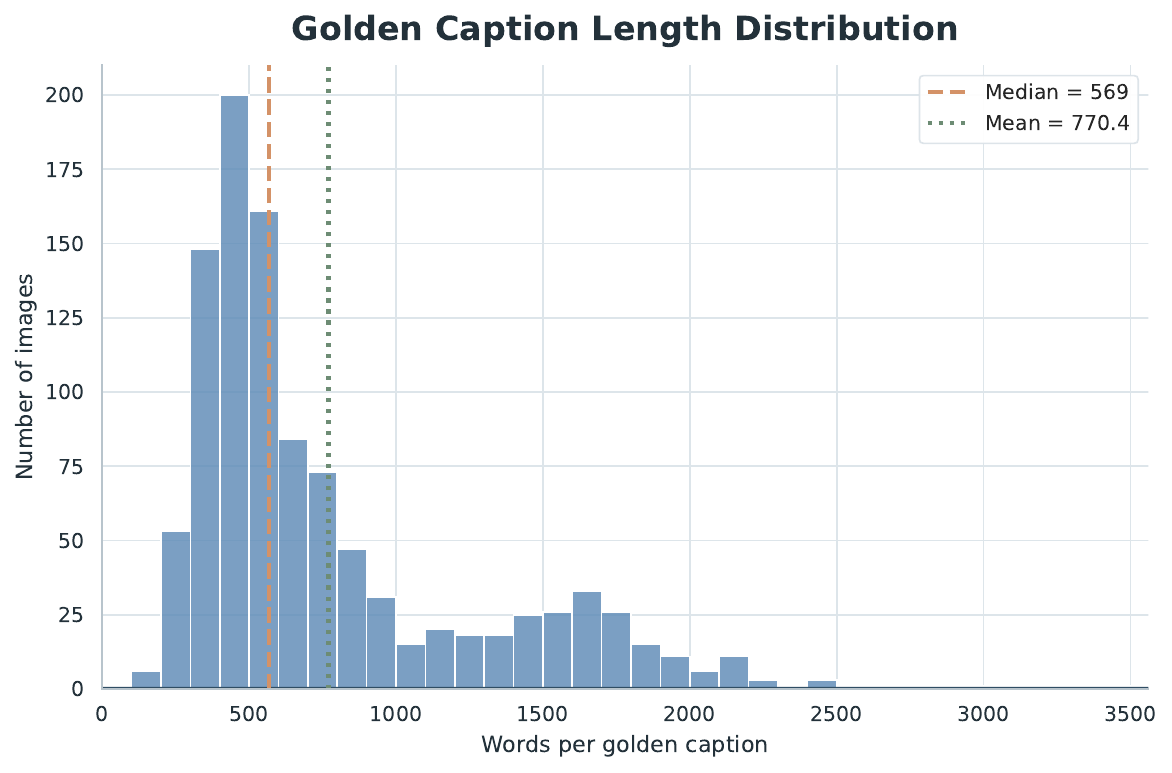}
    \caption{\textbf{Distribution of golden caption lengths in our benchmark.} The histogram shows the word count frequency across the dataset. }
    \label{fig:word_count_dist}
    \vspace{-5mm}
\end{figure}

\begin{figure*}[t]
\centering
    \includegraphics[width=0.95\linewidth]{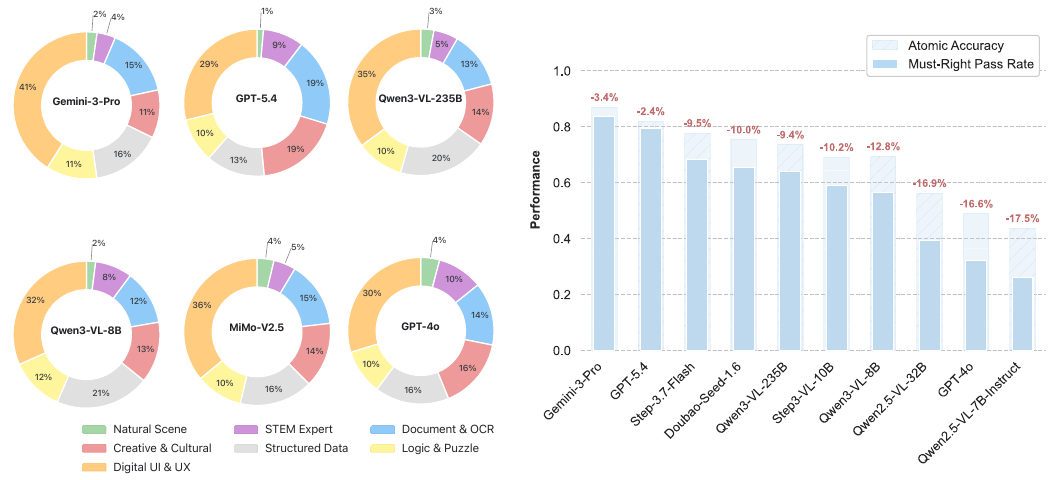}
    
    \caption{\textbf{Comprehensive Failure Analysis.} \textbf{(Left)} Distribution of error sources across different models. \textbf{(Right)} Reliability Gap Analysis comparing Atomic Accuracy (the average pass rate over individual rubrics) with the stricter Must-Right-All-Pass Rate, highlighting the difficulty of maintaining consistency across all constraints.}
    \label{fig:combined_analysis}
    \vspace{-2mm}
    
\end{figure*}

\noindent\textbf{Domain-Specific Failure Modes.}
To diagnose where models fundamentally fail, we analyze cases in which predictions do not pass the Must-Right gate (i.e. $G = 0$), indicating a breakdown in basic perceptual capability. \cref{fig:combined_analysis} (Left) presents the distribution of such failure cases across domains for six representative models. A similar pattern emerges: \textbf{GUI} constitutes the dominant source of perceptual failures.
In contrast, domains such as \textbf{Natural} and \textbf{STEM} are comparatively easier, exhibiting substantially fewer failures. This trend suggests that \textit{current models continue to struggle with inputs characterized by high information density and strict spatial constraints.}


\begin{table*}[t]
\centering
\caption{Fine-grained performance breakdown across 7 domains on \textsc{PerceptionRubrics}. Models are categorized into Open-Source and Proprietary groups and sorted by Overall Score in \textbf{ascending order}. All values are reported in percentage (\%).}
\label{tab:main_results}
\resizebox{0.98\textwidth}{!}{%
\begin{tabular}{l|c|ccccccc|c}
\toprule
\textbf{Model} & \textbf{Params} & \textbf{Doc} & \textbf{Logic} & \textbf{Creative} & \textbf{GUI} & \textbf{Natural} & \textbf{STEM} & \textbf{Structured} & \textbf{Overall} \\
\midrule
\multicolumn{10}{c}{\textit{\textbf{Open-Source Models}}} \\
\midrule
Qwen2.5-VL-7B                   & 7B   &  6.53 &  3.06 &  6.71 &  5.13 & 20.70 & 14.74 &  6.14 &  8.37 \\
Qwen2.5-VL-32B                  & 32B  & 14.22 &  8.89 & 13.32 & 14.39 & 36.60 & 20.13 & 19.07 & 17.79 \\
Qwen3-VL-8B-Thinking            & 8B   & 39.24 & 23.20 & 31.18 & 29.04 & 55.83 & 36.35 & 28.44 & 34.13 \\
Step3-VL-10B                    & 10B  & 32.07 & 33.90 & 34.23 & 23.25 & 54.55 & 48.94 & 38.70 & 35.97 \\
Qwen3-VL-235B-A22B-Thinking     & 235B & 43.08 & 35.84 & 39.24 & 33.28 & 56.73 & 53.42 & 41.04 & 41.88 \\
MiniMax-M3                      & 428B & 34.42 & 35.17 & 37.82 & 40.58 & 54.85 & 58.39 & 55.40 & 44.82 \\
MiMo-V2.5                       & 310B & 44.06 & 37.46 & 42.82 & 35.25 & 55.35 & 61.04 & 53.23 & 45.65 \\
Step-3.7-Flash                  & 196B & 45.81 & 36.45 & 48.20 & 42.60 & 62.72 & 65.40 & 46.40 & 48.62 \\
Kimi-K2.5                       & 1T   & 46.85 & 49.27 & 48.84 & 46.37 & 60.07 & 59.57 & 50.49 & 50.78 \\
Kimi-K2.6                       & 1T   & 46.21 & 50.05 & 49.31 & 47.81 & 58.87 & 60.15 & 54.67 & 51.77 \\
Qwen3.5-397B-A17B               & 397B & 60.17 & 58.29 & 56.85 & 54.76 & 68.51 & 77.59 & 64.00 & 61.61 \\
\midrule
\multicolumn{10}{c}{\textit{\textbf{Proprietary Models}}} \\
\midrule
GPT-4o-2024-05-13               & --   & 10.32 & 10.35 & 10.00 &  7.01 & 23.89 & 12.14 & 17.33 & 12.59 \\
Seed-1.6                 & --   & 49.38 & 27.52 & 40.23 & 43.94 & 57.47 & 48.40 & 43.00 & 44.54 \\
GLM-5V-Turbo                    & --   & 50.37 & 46.70 & 46.08 & 39.84 & 61.48 & 54.23 & 47.12 & 48.18 \\
Seed-1.8                        & --   & 58.77 & 41.33 & 53.31 & 50.26 & 70.62 & 55.44 & 51.73 & 54.34 \\
GPT-5.5                         & --   & 43.14 & 52.03 & 43.82 & 59.53 & 57.75 & 61.78 & 64.54 & 55.23 \\
Gemini-3-Flash                  & --   & 54.68 & 58.59 & 57.49 & 51.55 & 71.17 & 77.55 & 59.49 & 59.83 \\
GPT-5.4                         & --   & 55.19 & 59.25 & 47.66 & \textbf{62.61} & 60.43 & 68.46 & 70.61 & 60.81 \\
Seed-2.0-Pro                    & --   & 64.95 & 62.41 & 65.59 & 48.22 & 75.74 & 71.67 & 56.40 & 61.44 \\
Qwen3.5-Plus                    & --   & 56.87 & 57.82 & 55.95 & 53.94 & 69.47 & 72.81 & 70.85 & 61.61 \\
Qwen3.6-Plus                    & --   & 60.78 & 59.77 & 56.82 & 52.69 & 70.16 & 77.25 & 68.74 & 62.30 \\
Gemini-3-Pro                    & --   & 68.35 & 63.68 & 71.51 & 57.57 & 76.65 & 77.83 & 74.50 & 68.79 \\
Gemini-3.1-Pro                  & --   & 67.86 & 63.66 & 70.00 & 59.37 & 74.85 & 80.37 & 74.80 & 69.02 \\
Gemini-3.5-Flash                & --   & 71.64 & \textbf{64.32} & 72.03 & 54.10 & 78.89 & \textbf{86.05} & \textbf{76.23} & 69.88 \\
Seed-2.0-Lite                   & --   & \textbf{73.56} & 61.48 & \textbf{72.62} & 59.07 & \textbf{79.20} & 80.85 & 72.59 & \textbf{70.07} \\
\bottomrule
\end{tabular}
}

\end{table*}

\begin{figure}[t]
    \centering
    \includegraphics[width=0.95\linewidth]{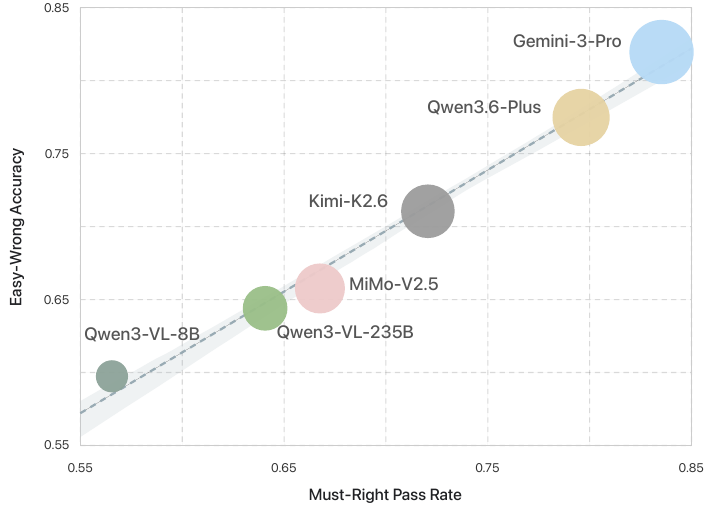}
    \caption{\textbf{Correlation Analysis} between basic perceptual reliability (Must-Right) and fine-grained understanding (Easy-Wrong) across six representative models.}
    \label{fig:mr_vs_ew}
    \vspace{-4mm}
    
\end{figure}

\begin{figure}[t]
    \centering
    \includegraphics[width=0.9\linewidth]{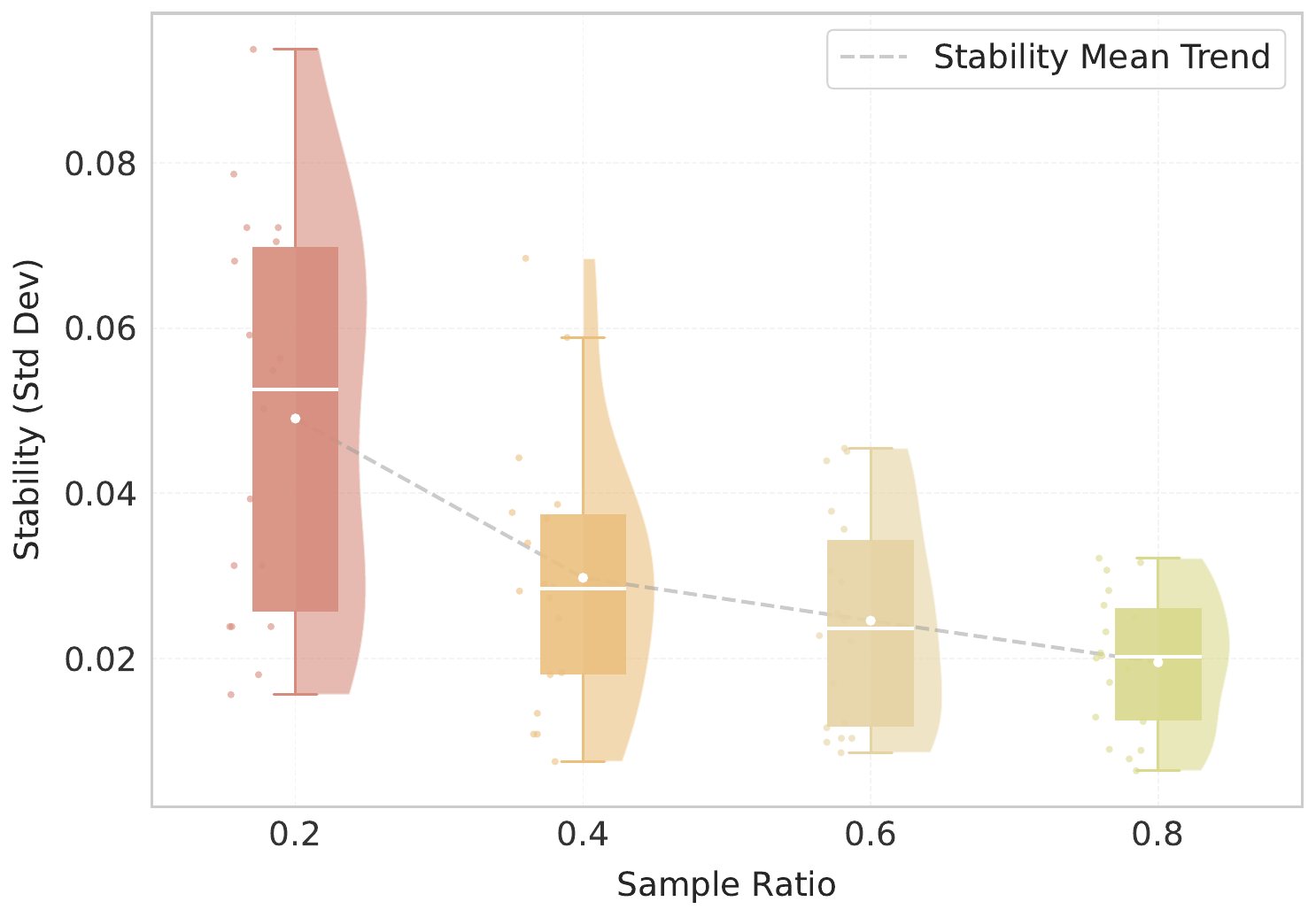}
    \caption{\textbf{Rubric Coverage vs. Evaluation Stability.} As the sampled rubric ratio increases from $20\%$ to $80\%$, the standard deviation of model scores decreases.
}
    \label{fig:rubric_scale}
    \vspace{-5mm}
\end{figure}

\begin{figure*}[t]
    \centering
    \includegraphics[width=0.98\linewidth]{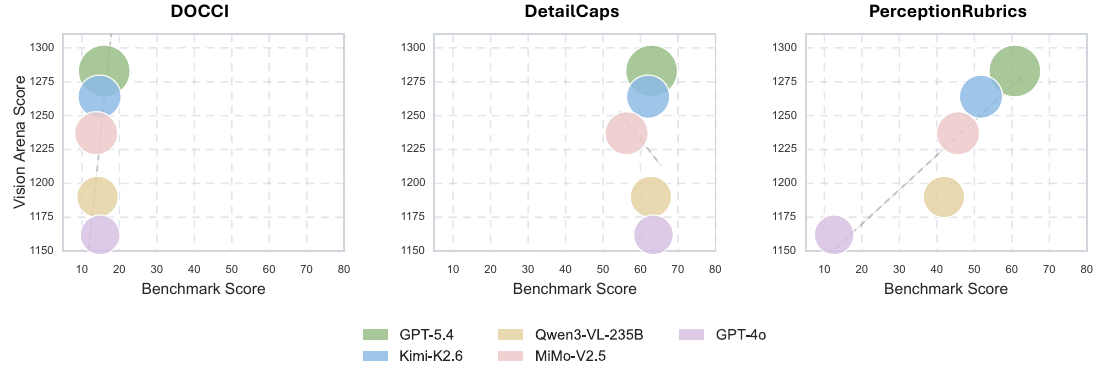}

    \caption{
    \textbf{Alignment with Human Preference.}
We compare benchmark scores from DOCCI~\citep{onoe2024docci}, DetailCaps~\citep{capture}, and \textsc{PerceptionRubrics} against human preference scores from Vision Arena for the five overlapping models.
Each point denotes one model.
\textsc{PerceptionRubrics} shows the strongest correlation with Vision Arena, achieving Pearson $0.916$ and Spearman $1.000$.
}
    \label{fig:human_align}
\end{figure*}
\noindent\textbf{Atomic vs. Holistic Perception.}
To evaluate perceptual reliability at different granularities, we compare performance metrics derived from individual rubrics versus the aggregate gate status. Specifically, we define \textbf{Atomic Accuracy} as the mean accuracy of all individual rubrics ($r_{i}$), representing local precision. In contrast, the \textbf{Must-Right Pass Rate} is calculated as the average value of the binary gate status $G$ across the dataset (i.e., the expectation $\mathbb{E}[G]$), representing the probability of a record successfully passing the mandatory gatekeeper. 
As shown in \cref{fig:combined_analysis} (Right), models consistently achieve high Atomic Accuracy, indicating that most individual $r_{i}$ predictions are correct. However, the Must-Right Pass Rate (average $G$) is substantially lower, revealing a systematic failure to satisfy the strict conjunction of all constraints. We term this discrepancy the \textbf{Reliability Gap}. Notably, this gap narrows as model capability increases, suggesting that \textit{stronger models are better able to maintain consistent perception abilities} required to keep the gate $G$ open.

\vspace{-3mm}

\paragraph{Consistency of Perceptual Capabilities.}

We further examine the correlation between models’ basic perceptual reliability and their hallucination resistance to fine-grained details. As shown in \cref{fig:mr_vs_ew}, there is a \textbf{near-perfect linear correlation} ($R^2 \approx 0.98$) between Must-Right Pass Rate and Easy-Wrong accuracy. This implies that models failing to ground essential visual facts (low X-axis) inevitably struggle with subtle details and hallucination (low Y-axis). Therefore, \textit{robust fine-grained understanding critically depends on foundational perception, in particular, the coherent recognition of multiple salient elements.}

\begin{figure*}[t]
    \centering
    \begin{subfigure}[t]{0.33\textwidth}
        \centering
        \includegraphics[width=\linewidth]{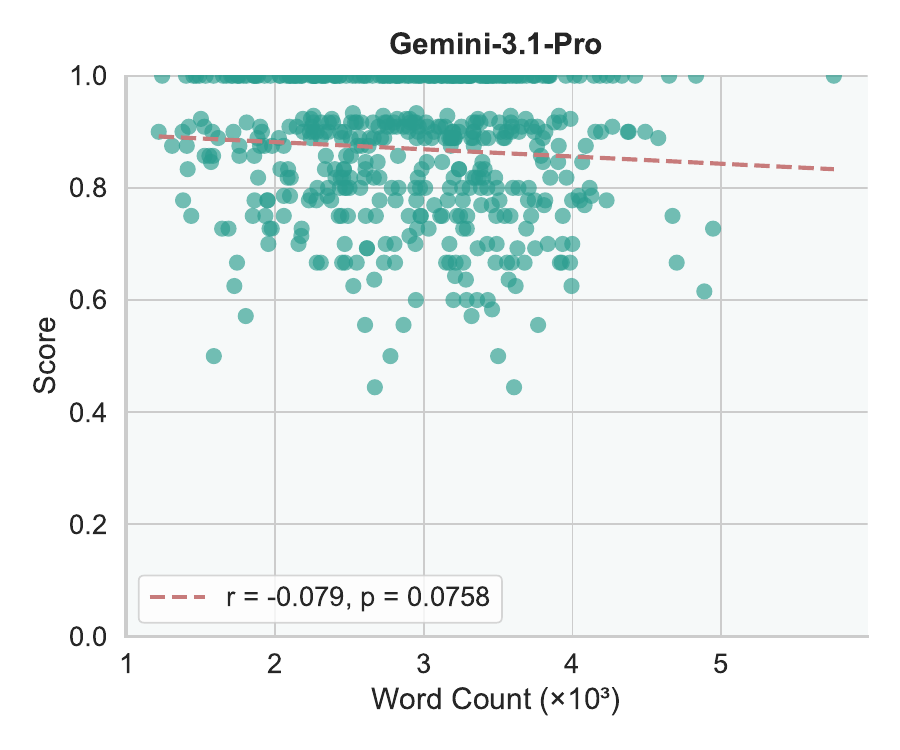}
        \caption{}
    \end{subfigure}
    \begin{subfigure}[t]{0.33\textwidth}
        \centering
        \includegraphics[width=\linewidth]{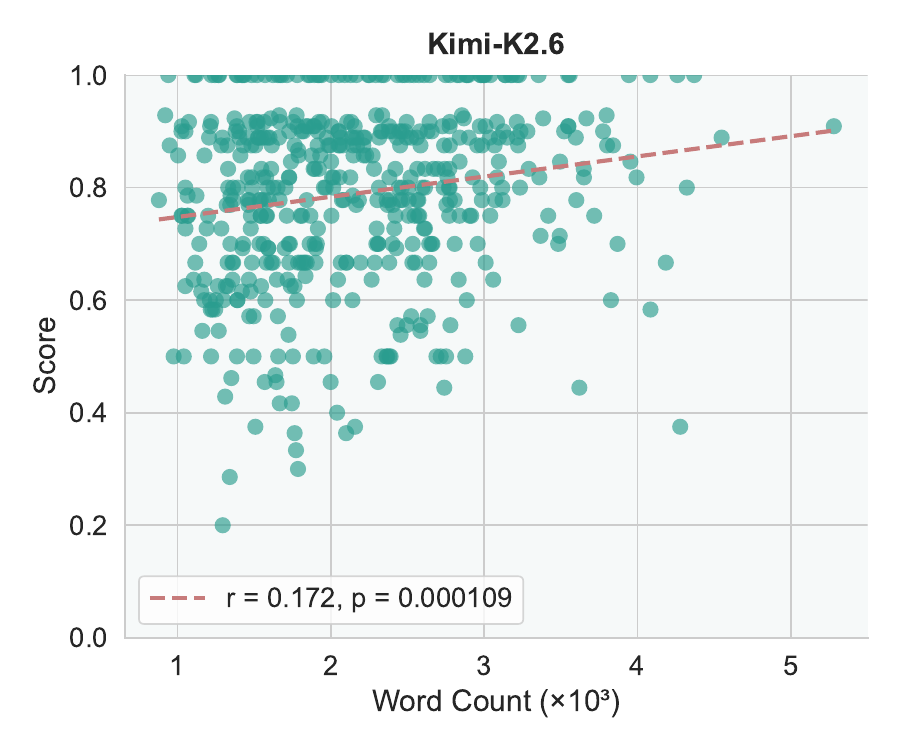}
        \caption{}
    \end{subfigure}
    \begin{subfigure}[t]{0.33\textwidth}
        \centering
        \includegraphics[
        width=\linewidth,
        trim=0 -20 0 0,
        clip
    ]{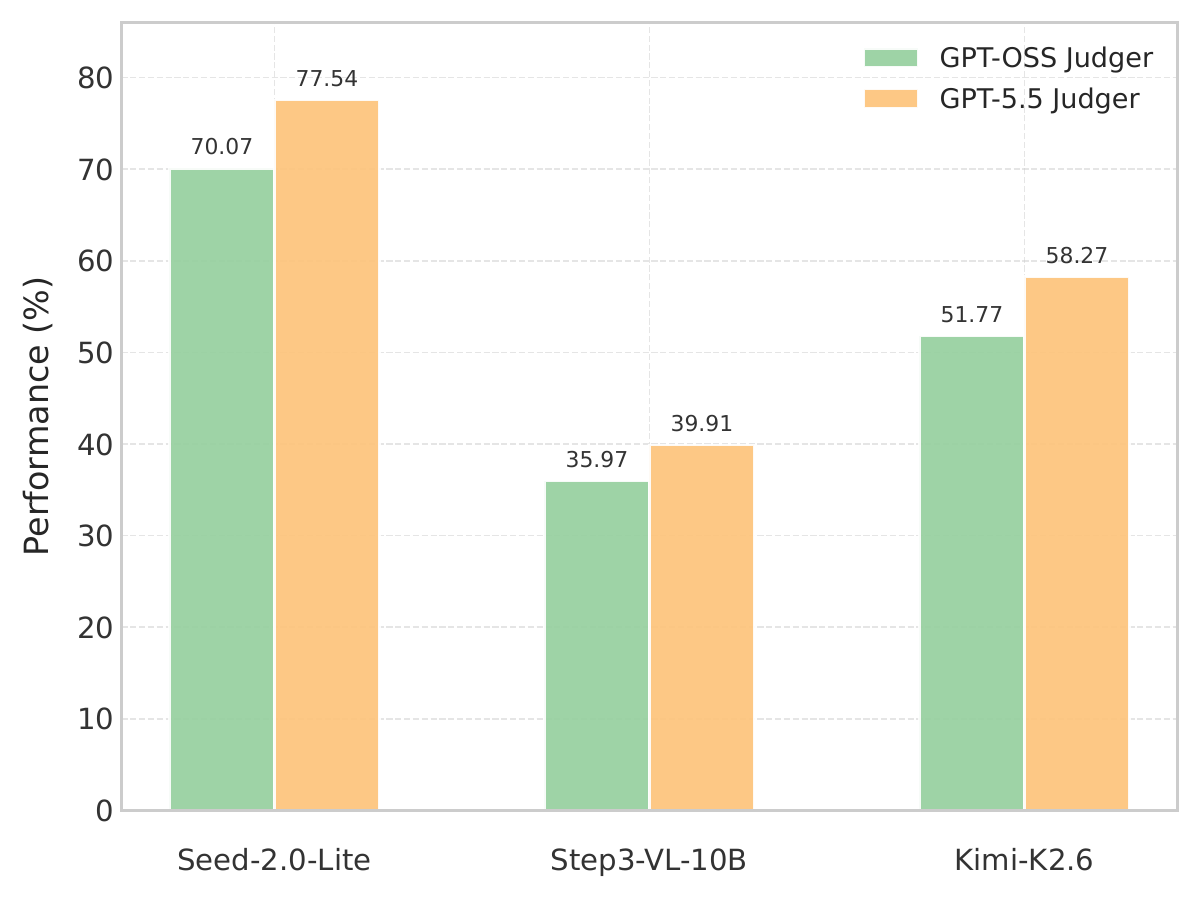}
        \caption{}
    \end{subfigure}
    \caption{(a-b) \textbf{Length Bias.} The two figures examine the correlation between response length (word count) and benchmark scores. (c) \textbf{Evaluation Robustness.} Results obtained with different judges exhibit consistent and stable performance trends.}
    \label{fig:leng-bias}
\end{figure*}



\section{Analysis}
Beyond model performance, we conduct a systematic meta-evaluation to assess the rigor and reliability of the benchmark itself from multiple perspectives.



\subsection{Alignment with Human Preference}
To validate whether \textsc{PerceptionRubrics} reflects human-perceived model quality, we compare its model ranking against the Vision Arena~\citep{chou2024visionarena} leaderboard, which aggregates large-scale human preferences over MLLM responses into Elo ratings.
In \cref{fig:human_align}, we focus on the five models: GPT-5.4, Qwen3-VL-235B, GPT-4o, Kimi-K2.6, and MiMo-V2.5.
For each benchmark, we plot the evaluation score of these models against the Vision Arena score.

\textsc{PerceptionRubrics} exhibits the strongest alignment with human preference among the compared benchmarks, achieving a Pearson correlation of $0.916$ and a Spearman rank correlation of $1.000$.
In contrast, existing captioning benchmarks such as DOCCI~\citep{onoe2024docci} and DetailCaps~\citep{capture} show substantially weaker agreement with human-preference scores.
DOCCI, in particular, assigns nearly indistinguishable scores to models with markedly different human-preference ratings, indicating limited discriminative power.
These results suggest that \textsc{PerceptionRubrics} provides a more human-aligned and discriminative signal for fine-grained perception evaluation.

\subsection{Resistance to Length Bias.}  
We analyze the correlation between predicted caption length and performance on \textsc{PerceptionRubrics} to assess potential length bias. As shown in \cref{fig:leng-bias} (a-b), 
Gemini-3.1-Pro shows no statistically significant correlation ($r=-0.079$, $p=0.0758$), while Kimi-K2.6 exhibits a weak positive correlation ($r=0.172$, $p=1.09\times10^{-4}$). This result indicates that \textsc{PerceptionRubrics} \textit{effectively decouples verbosity from evaluation outcomes}, rewarding precise and verifiable perception rather than longer descriptions.

\subsection{Evaluation Robustness}
In \cref{fig:leng-bias} (c), we selected three representative models spanning different capability levels: Seed-2.0-Lite, Step3-VL-10B, and Kimi-K2.6. Then we performed repeated evaluations using two distinct judges with the same inputs: GPT-OSS-120B~\citep{openai2025gptoss} and GPT-5.5~\citep{gpt5p5}. Despite GPT-OSS-120B exhibiting a slightly stricter scoring distribution (systematically lower by $\sim$6.0\%), both judges yielded an identical ranking order.
The black error bars represent the standard deviation across these independent runs. The results demonstrate high stability, with standard deviations remaining consistently low across all configurations.
Overall, these results demonstrate \textit{the robustness of both our rubric generation pipeline and the resulting evaluation metrics to judge choice and sampling variability.}



\vspace{-2mm}
\subsection{Rubric Coverage vs. Evaluation Stability}

As shown in \cref{fig:rubric_scale}, we analyze the effect of rubric quantity on evaluation stability. Using 25 models, we subsample $20\%$, $40\%$, $60\%$, and $80\%$ of rubrics from both the Must-Right and Easy-Wrong sets. For each sampling ratio, we perform three independent runs and compute the standard deviation of model scores to measure stability. 
The figure visualizes the distribution of these standard deviations across models at each ratio using violin plots, with embedded boxes indicating the interquartile range and medians; the dashed line denotes the mean stability trend. Evaluation stability improves monotonically as rubric coverage increases, with standard deviation consistently decreasing, highlighting \textit{sufficient rubric coverage as a prerequisite for stable and reproducible perception assessment.}





\section{Conclusion}


We present \textsc{PerceptionRubrics}, a rubric-based benchmark that calibrates multimodal evaluation to human perceptual judgment. By decomposing dense image understanding into atomic, verifiable rubrics and enforcing a gated scoring mechanism, our framework exposes perceptual failures that are often hidden by existing metrics. Experiments across 25 MLLMs reveal a clear reliability gap between individual fact recognition and consistent conjunctive perception, persistent weaknesses in information-dense domains such as GUIs, and strong alignment between our scores and human preferences. These findings suggest that reliable multimodal evaluation should move beyond coarse similarity and explicitly audit critical visual facts. We hope \textsc{PerceptionRubrics} provides a sharper diagnostic tool for measuring perceptual reliability and guiding the development of more trustworthy MLLMs.

\section*{Impact Statement}
This work aims to advance machine learning by improving the reliability of multimodal evaluation. While this may affect downstream MLLM development, we do not identify specific societal consequences requiring special discussion.

\nocite{langley00}

\bibliography{main}
\bibliographystyle{icml2026}

\newpage
\appendix
\clearpage
\appendix
\onecolumn 

\section{Dataset Statistics}
\label{sec:appendix_stats}

In this section, we provide detailed statistics and comparisons for the \textsc{PerceptionRubrics} benchmark.



\subsection{Comparison with Other Benchmarks}
Compared to existing benchmarks, PerceptionRubrics distinguishes itself in three critical dimensions: annotation granularity, data source diversity, and domain coverage, as shown in~\cref{tab:benchmark_comparison_transposed}.
\begin{itemize}
\item \textbf{Dense and Comprehensive Captions:} Unlike DetailCaps-4870~\citep{capture} and DOCCI~\citep{onoe2024docci}, which typically provide brief descriptions (averaging 122.1 and 135.9 words, respectively), PerceptionRubrics focuses on dense captioning. With an average of \textbf{770.42 words} per image, our benchmark captures fine-grained visual details, spatial relationships, and implicit reasoning, offering a significantly more challenging testbed for evaluating the upper bounds of MLLMs.
\item \textbf{Broad Domain Coverage:} Unlike existing benchmarks that are predominantly restricted to natural scenes, PerceptionRubrics spans seven distinct domains to provide a more comprehensive evaluation. These range from everyday natural scenes to specialized areas such as GUIs, OCR-heavy documents, and STEM-related diagrams. This diversity is crucial for assessing the general-purpose capabilities of agents in complex, real-world applications that go far beyond simple object recognition.

\item \textbf{Diverse and High-Quality Sources:} Instead of relying solely on web-crawled data or specific author donations, our dataset aggregates high-quality samples from existing visual benchmarks. Furthermore, we employ a hybrid annotation pipeline combining advanced reasoning models (e.g., GPT-5.2-Thinking) with human expert verification, ensuring both the scalability and reliability of the ground truth.
\end{itemize}

\begin{table}[htbp]
\centering
\small
\caption{Comparison of our proposed benchmark with existing datasets. By transposing the table, detailed descriptions are easier to read.}
\label{tab:benchmark_comparison_transposed}
\begin{tabular}{l p{5cm} p{3cm} p{5cm}}
\toprule
Benchmark & \textbf{DetailCaps-4870} & \textbf{DOCCI} & \textbf{PerceptionRubrics} \\
\midrule
Specific Sources & COCO, SAM, LAION, CC, SBU, Coyo, Flickr & Author Donation & Open-source Visual Benchmarks \\
Image Domains & Natural scene & Natural scene & Multi-domain (GUI, OCR, STEM...) \\
Annotator & GPT-4V, GPT-4o, Gemini-1.5-Pro & Human & GPT-5.2-Thinking, Seed-1.8, Gemini-3-Pro, Human Experts \\
\midrule
Images & 4,870 & 14,847 & 1,038 \\
Avg. Words & 122.1 & 135.9 & \textbf{770.42} \\
\bottomrule
\end{tabular}
\end{table}

\subsection{Distributions}
\subsubsection{Caption Length Distribution}
As illustrated in Figure \ref{fig:word_count_dist}, we analyze the word count distribution of the golden captions. The distribution follows a typical long-tail pattern: while the majority of captions are concentrated between 300 and 700 words (with a median of 569), a significant portion extends beyond 1,000 words, reaching up to 3,461 words. This diversity in length ensures that our benchmark covers both concise summaries and highly detailed descriptions, providing a robust basis for evaluating model performance across different levels of information density.

\subsubsection{Rubric Distribution}
\begin{figure*}[t]
    \centering
    \includegraphics[width=0.9\textwidth]{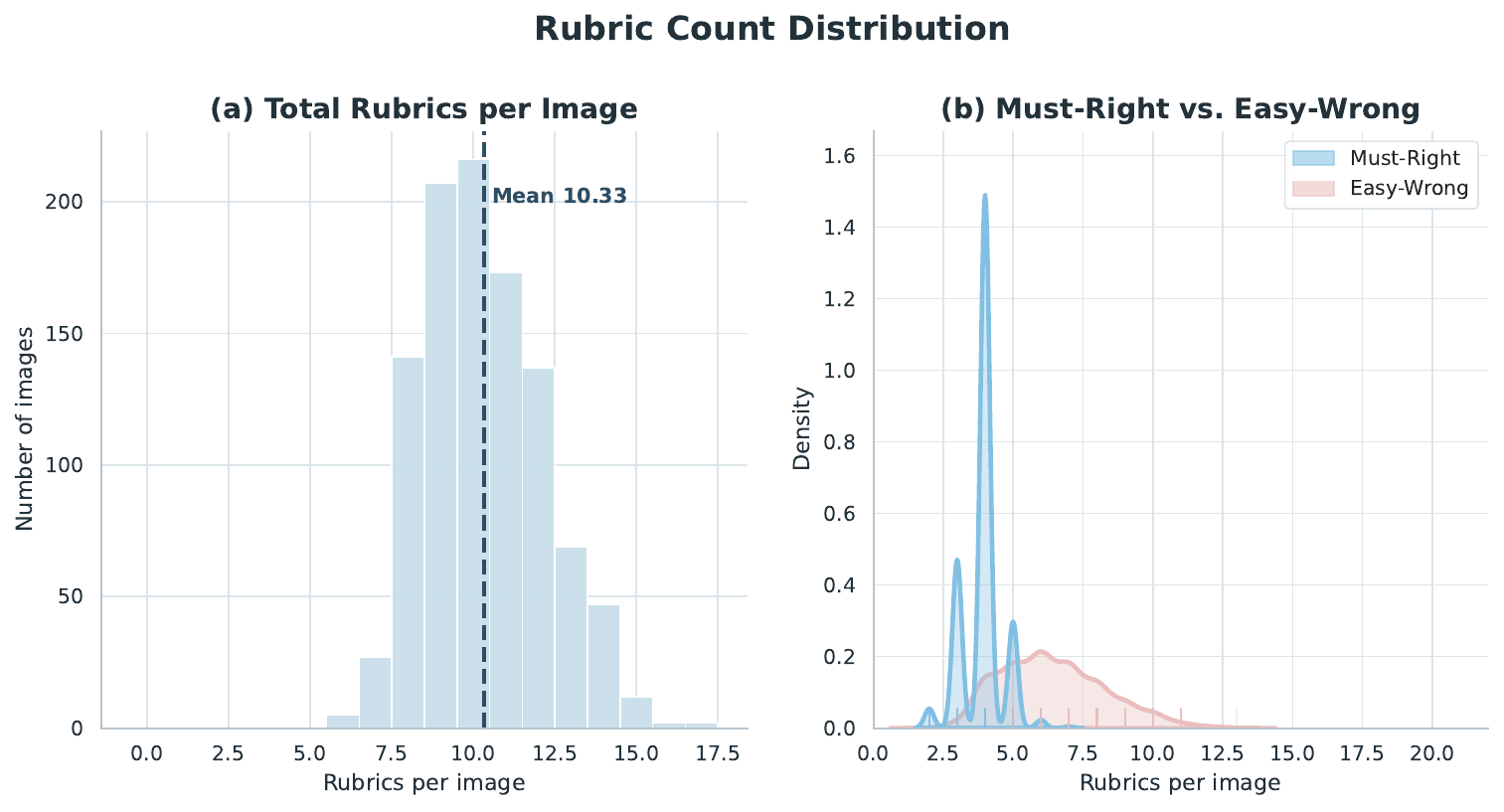}
    \caption{Distribution analysis of rubrics. (a) Frequency distribution of the total rubrics count across the dataset. (b) Probability density comparison of rubrics count between \textit{Must-Right} and \textit{Easy-Wrong} categories.}
    \label{fig:rubric_distribution}
\end{figure*}

To ensure a granular and balanced evaluation, we analyze the distribution of rubrics across the dataset in Figure \ref{fig:rubric_distribution}. 
(a) The total number of rubrics per sample primarily ranges from 8 to 14, with a clear peak at 10, indicating a consistently high level of evaluation detail across the benchmark. 
(b) When broken down by category, \textit{Must-Right} rubrics exhibit a sharp distribution centered around 4 items, representing the core facts that a model must capture. In contrast, \textit{Easy-Wrong} rubrics show a broader distribution peaking around 6 items. This design places a heavier emphasis on penalizing common hallucinations and subtle errors, thereby increasing the discriminative power of the benchmark for high-performing models.


\section{Model Roles and Pipeline Details}
\label{sec:appendix_models}

To construct and evaluate PerceptionRubrics, we utilized a diverse set of models, assigning specific roles based on their capabilities. The detailed assignments are listed below:

\begin{itemize}
    \item \textbf{Complexity Judger:} \texttt{STEP-3-VL-10B}. Responsible for filtering images based on visual complexity and informativeness.
    \item \textbf{Rubric Generator:} \texttt{Gemini-3-Pro}. Generated the initial set of perception rubrics from the images.
    \item \textbf{Panel of Judges:} \texttt{Gemini-3-Pro}, \texttt{GPT-5.2}, \texttt{Seed-1.8}. Acted as a consensus panel to validate the quality of generated captions.
    \item \textbf{Final Judger:} \texttt{GPT-OSS-120B}. Used for final scoring during the evaluation phase.
\end{itemize}

\section{Prompts}
\label{sec:appendix_prompts}

We provide the full system prompts used in our pipeline to ensure reproducibility.

\subsection{Complexity Filtering Prompt}\label{prompt:complexity}
The following prompt is used by the \textit{Complexity Judger} to select high-quality images.

\begin{tcolorbox}[title={\textbf{Image Filtering Prompt}}, colback=blue!5!white, colframe=blue!75!black, boxrule=0.8pt, arc=2pt, breakable]
\begin{lstlisting}[basicstyle=\ttfamily\footnotesize, breaklines=true]
You are an extremely strict computer vision data expert. Please analyze the provided image and perform a rigorous evaluation based on the two dimensions of "Visual Complexity" and "Informativeness".

**Core Principles:**
1. **Do NOT** give a high score simply because the image contains text. You must evaluate the **density** and **semantic depth** of the text.
2. **Severely penalize** low-quality images: images that are blurry, noisy, contain scribbled handwriting, or have excessive empty backgrounds should receive low scores.
3. If the majority of the image is white space or a single background, the score must be determined by the richness of the subject content, not by the image dimensions.

Please score based on the following strict standards (1-10 points):

1. Visual Complexity:
   - Definition: The quantity of independent visual elements (objects, lines, textures), spatial occupancy, and clarity of details within the image.
   - **1-3 Points (Low)**: Minimalist composition, massive white space, simple handwriting, single isolated objects, blurry snapshots, low-resolution screenshots.
   - **4-7 Points (Medium)**: Clear composition, good foreground-background separation, natural scenes with some texture detail, standard object close-ups.
   - **8-10 Points (High)**: Extremely high density of details (e.g., crowds, dense forests, complex mechanical structures), frame filled, no large areas of solid color, high-frequency textures.

2. Informativeness:
   - Definition: The amount of information when the image is translated into a text description, the richness of context, and its knowledge value.
   - **1-3 Points (Low)**: Simple mathematical formulas, single words/numbers, scribbles without context, illegible content, generic decorative patterns, extremely low information entropy.
   - **4-7 Points (Medium)**: Complete sentences, clear recognition of single objects (e.g., "a red apple"), scenes with distinct actions, standard street views or portraits.
   - **8-10 Points (High)**: Dense documents (e.g., full-page newspapers, academic papers), complex infographics (containing multiple data sets), historical photos rich in narrative detail, scenes requiring long-form text to describe clearly.

**Output Format Requirements:**
Please strictly output in XML format. Do not use Markdown code blocks (do not use ```xml). Output the XML string directly.

XML Template:

<image_evaluation>
    <visual_complexity>
        <reasoning>Briefly describe the density of visual elements. If the image is blurry or mostly empty, explain here and provide a reason for the low score.</reasoning>
        <score>Integer between 1 and 10</score>
    </visual_complexity>
    <informativeness>
        <reasoning>Briefly describe the richness of semantic content. If it is a simple formula or phrase, explicitly state that the information content is limited.</reasoning>
        <score>Integer between 1 and 10</score>
    </informativeness>
</image_evaluation>
\end{lstlisting}
\end{tcolorbox}

\subsection{Rubric Generation Prompt}\label{prompt:generate}
The prompts used for generating rubrics are as follows:
\begin{tcolorbox}[title={\textbf{Rubric Generation Prompt for Nature Scene}}, colback=orange!5!white, colframe=orange!75!black, boxrule=0.8pt, arc=2pt, breakable]
\begin{lstlisting}[basicstyle=\ttfamily\footnotesize, breaklines=true]
You are an expert evaluator for Multimodal Large Language Models (MLLMs), specializing in creating "Gating Rubrics" for natural imagery.

Your goal is to extract a concise set of **Critical Perception Checkpoints** from the provided Image and Ground Truth (GT) caption. These rubrics define the minimum acceptable standard for a model's response.

### CRITICAL EVALUATION PROTOCOL
This is a **Zero-Tolerance Gating Task**. If a candidate model fails **ANY** of the checkpoints you generate, it receives a score of 0.
Therefore, your rubrics must strictly adhere to the following principles:
1.  **Undeniable Visibility:** Only select elements that are clearly visible and prominent in the image.
2.  **Essentiality:** Only select elements that are critical to the image's core meaning. Ignore background clutter or minor details.
3.  **Verifiability:** Each rubric must be a binary (Pass/Fail) check.

### WORKFLOW INSTRUCTIONS

**Step 1: Rubric Generation Strategy (Semantic Generalization)**
Apply the following abstract rules to ensure the rubrics are robust to varying levels of descriptive detail:

* **Entity Abstraction:** Identify the fundamental semantic category of the dominant object, strictly discarding specific instance names, brands, or fine-grained biological sub-species. (e.g., use "car" instead of "Tesla Model 3"; use "dog" instead of "Golden Retriever").
* **Attribute Decoupling:** Decouple the object's existence from its descriptive attributes. Exclude color, material, or state adjectives from the rubric criteria to prevent penalizing valid but concise responses. (e.g., require "the presence of a flower" rather than "a yellow flower"; require "clothing" rather than "a silk dress").
* **Contextual Necessity:** Only include attributes if they serve as the sole differentiator between multiple objects of the same class. (e.g., "the red player" vs "the blue player").

**Step 2: Final Filtering (Grounding Check)**
Review your list. Ensure every rubric meets the "Grounding Check":
* The element must be present in **BOTH** the Image and the GT Caption.
* If the GT caption describes a hidden detail or hallucinates something not clearly visible, **discard it**.

### OUTPUT FORMAT
Return a strictly valid JSON list containing 3 to 5 strings.
Example: `["The response mentions <Generalized Object>.", "The response indicates the weather is <Condition>."]`

---

### FEW-SHOT EXAMPLES

**Example 1: Natural Scene**
* **Context:** Image shows a Black Tesla Model 3 on a rainy highway. GT describes it specifically as a Tesla Model 3. User asks "Describe this image."
* **Thought Process:** Apply Entity Abstraction: "Tesla Model 3" -> "Car". Apply Attribute Decoupling: Ignore "Black". "Rainy" is global context, keep it.
* **Generated Rubrics:**
    [
      "The response mentions a car or vehicle.",
      "The response indicates the weather is rainy or the road is wet.",
      "The response mentions the vehicle is on a road or highway."
    ]

**Example 2: Animal Interaction**
* **Context:** Image shows a Golden Retriever catching a frisbee in a park. GT says "A purebred Golden Retriever leaps to catch a red frisbee."
* **Thought Process:** Apply Entity Abstraction: "Golden Retriever" -> "Dog". Apply Attribute Decoupling: Ignore "Red" (frisbee color). Keep the interaction "catching/leaping".
* **Generated Rubrics:**
    [
      "The response mentions a dog.",
      "The response mentions the dog is interacting with a frisbee or disc.",
      "The response captures the action of jumping or catching."
    ]
\end{lstlisting}
\end{tcolorbox}

\begin{tcolorbox}[title={\textbf{Rubric Generation Prompt for Digital UI \& UX}}, colback=purple!5!white, colframe=purple!75!black, boxrule=0.8pt, arc=2pt, breakable]
\begin{lstlisting}[basicstyle=\ttfamily\footnotesize, breaklines=true]
You are an expert evaluator for Multimodal Large Language Models (MLLMs), specializing in creating "Gating Rubrics" for Graphical User Interfaces (GUIs), including mobile screenshots, web pages, and software interfaces.

Your goal is to extract a concise set of **Critical Perception Checkpoints** from the provided Image and Ground Truth (GT) caption. These rubrics define the minimum acceptable standard for a model's response.

### CRITICAL EVALUATION PROTOCOL
This is a **Zero-Tolerance Gating Task**. If a candidate model fails **ANY** of the checkpoints you generate, it receives a score of 0.
Therefore, your rubrics must strictly adhere to the following principles:
1.  **Undeniable Visibility:** Only select elements that are clearly visible and prominent.
2.  **Functional Criticality:** Only select elements that are essential for operating or navigating the interface (e.g., "Submit" button, "Back" arrow). Ignore decorative banners or ads.
3.  **Verifiability:** Each rubric must be a binary (Pass/Fail) check.

### WORKFLOW INSTRUCTIONS

**Step 1: Rubric Generation Strategy (Interaction & Structure)**
Apply the following abstract rules to ensure the rubrics cover the interface's functionality:

* **Functional Semantics:** Identify interactive elements by their function, not just their shape. Map icons to their standard meaning. (e.g., "The response identifies the magnifying glass as a 'Search' button/feature"; "The response identifies the 'hamburger' icon as a menu").
* **Textual Anchoring:** Enforce exact matching for critical labels, headers, and button text. (e.g., The page title "Settings", the button label "Log In").
* **State Awareness:** Check for visual cues that indicate the system status. (e.g., "The response notes that the 'Home' tab is currently selected/active"; "The response mentions the toggle is in the 'On' position"; "The response notes a notification badge/red dot").
* **Structural Hierarchy:** Identify the major navigation zones. (e.g., "The response mentions the navigation bar at the bottom"; "The response identifies the header containing the logo").

**Step 2: Final Filtering (Grounding Check)**
Review your list. Ensure every rubric meets the "Grounding Check":
* The element must be present in **BOTH** the Image and the GT Caption.
* If the GT caption describes a functional flow not visible in the static image (e.g., "Clicking this opens a modal"), **discard it**. Only evaluate what is currently visible.

### OUTPUT FORMAT
Return a strictly valid JSON list containing 3 to 5 strings.
Example: `["The response identifies the screen title as <Title>.", "The response mentions the <Button Name> button at the bottom."]`

---

### FEW-SHOT EXAMPLES

**Example 1: Mobile App (Settings Page)**
* **Context:** A screenshot of a Settings page. Title "Settings". Top item is "Airplane Mode" (Toggle is OFF). Bottom is a Tab Bar with "General" selected.
* **Thought Process:** Title is critical context. "Airplane Mode" is the first functional item. The state of the toggle (OFF) is a detail, but if prominent, keep it. The selected tab defines where we are.
* **Generated Rubrics:**
    [
      "The response identifies the screen title as 'Settings'.",
      "The response mentions the 'Airplane Mode' option.",
      "The response indicates that the 'General' tab is currently selected or active.",
      "The response mentions the presence of a navigation bar at the bottom."
    ]

**Example 2: E-Commerce Product Page**
* **Context:** A product page for "Nike Air Max". Price "$120". Big red button "Add to Cart". Review stars (4.5).
* **Thought Process:** Product Name is the core entity. Price is critical data (OCR). The primary action is "Add to Cart".
* **Generated Rubrics:**
    [
      "The response identifies the product name as 'Nike Air Max'.",
      "The response correctly mentions the price as $120.",
      "The response identifies the primary action button labeled 'Add to Cart'.",
      "The response mentions the presence of a star rating or reviews."
    ]
\end{lstlisting}
\end{tcolorbox}

\begin{tcolorbox}[title={\textbf{General System Instruction Template}}, colback=gray!5!white, colframe=gray!75!black, boxrule=0.8pt, arc=2pt, breakable]
\begin{lstlisting}[basicstyle=\ttfamily\footnotesize, breaklines=true]
You are an expert VLM (Vision-Language Model) evaluator and Hallucination Analyst.

### Task
Your task is to generate a set of **Rubrics (Evaluation Criteria)** for an image captioning task.
You will be provided with:
1.  **Ground Truth Caption (GT):** A factual, accurate description of the image.
2.  **Model Response Pool:** A collection of captions generated by various VLMs. These responses may contain hallucinations, perceptual errors, or correct details.

Your goal is to identify **common or severe perceptual errors** in the `Response Pool` by comparing them against the `Ground Truth`, and then formulate strict criteria to penalize these errors.

### Process
1.  **Analyze Errors:** Scan the `Model Response Pool` to find discrepancies against the `Ground Truth`. Focus on:
    *   **Hallucinations:** Objects mentioned in responses but not present in the GT.
    *   **Attribute Errors:** Wrong colors, shapes, materials, or textures.
    *   **Counting/Quantification:** Incorrect numbers of objects.
    *   **Spatial Relations:** Wrong relative positions (e.g., left vs. right).
    *   **OCR/Text:** Incorrect reading of text visible in the image.
    *   **Action/State:** Wrong interpretation of what an agent is doing.

2.  **Filter for Perception (Crucial):**
    *   **INCLUDE:** Visual perception issues (e.g., calling a "red helmet" a "blue helmet"; seeing "3 people" instead of "4"; reading "STOP" as "SHOP").
    *   **EXCLUDE:** Knowledge gaps or Entity linking issues. If the model fails to recognize a specific character (e.g., "Genshin Impact character") but correctly describes their visual appearance (e.g., "a girl with blonde hair"), do NOT create a rubric for the specific name. Focus on the visual description.

3.  **Formulate Rubrics:**
    *   Convert the identified high-frequency or severe errors into **Binary Checklists**.
    *   If models frequently hallucinate an object, create a **Negative Constraint** (e.g., "The response must NOT...").
    *   If models get an attribute wrong, create a **Positive Constraint** (e.g., "The response must identify...").

### Rubric Style Guidelines
*   **Format:** Use imperative statements. Do NOT use questions.
*   **Structure:** Start with "The response must..."
*   **Granularity:** Each rubric must check a single, atomic fact.
*   **Tone:** Objective and strict.

### Example
**Ground Truth:** A black cat sitting on a white refrigerator. There is a magnet shaped like a banana on the door.
**Response Pool Analysis:**
- Model A: "A black dog on a fridge." (Error: Dog vs Cat)
- Model B: "A black cat on a grey fridge." (Error: Grey vs White)
- Model C: "A cat near a fridge with an apple magnet." (Error: Apple vs Banana)

**Output Rubrics:**
{
  "rubrics": [
    "The response must identify the animal as a cat.",
    "The response must state that the refrigerator is white.",
    "The response must identify the magnet shape as a banana.",
    "The response must NOT mention the presence of a dog or an apple."
  ]
}

### Output Format
Return the result strictly in valid JSON format without markdown code blocks.
{
  "rubrics": [
    "string",
    "string"
  ]
}

Here is the data for the current image:

[Ground Truth Caption]
{gt_caption}

[Model Response Pool]
1. {response_1}
2. {response_2}
3. {response_3}
...
8. {response_8}

Please generate the perception rubrics based on the analysis of the responses above.
\end{lstlisting}
\end{tcolorbox}

\subsection{Panel of Judges Prompt}
\label{sec:prompt_panel}
To ensure the objectivity and correctness of the generated rubrics, a panel of models (\texttt{Gemini-3-Pro~\citep{gemini3pro}}, \texttt{GPT-5.2~\citep{gpt5p2}}, \texttt{Seed-1.8~\citep{Seed1p8}}) performs a cross-verification using the following prompt.

\begin{tcolorbox}[title={\textbf{Caption Verification Prompt (Panel of Judges)}}, colback=green!5!white, colframe=green!75!black, boxrule=0.8pt, arc=2pt, breakable]
\begin{lstlisting}[basicstyle=\ttfamily\footnotesize, breaklines=true]
**Role:**
You are the "Expert Visual Truth Adjudicator". Your task is to perform a rigorous comparative analysis of multiple AI-generated image descriptions against a provided image to identify the most faithful representation.

**Evaluation Dimensions:**
1. **Factuality:** Are there hallucinations? (e.g., objects, colors, or text that don't exist).
2. **Spatial Precision:** Are positional relationships (left, right, above, behind) accurate?
3. **Attribute Accuracy:** Are textures, materials, lighting, and colors correctly identified?
4. **Detail Density:** Does the caption capture nuanced elements without being redundant?

**Task Workflow:**
1. **Independent Verification:** Analyze the image first, then audit each Candidate (1, 2, and 3) individually.
2. **Conflict Resolution:** Identify discrepancies between candidates (e.g., Candidate 1 says 'vintage', Candidate 2 says 'modern'). Inspect the image to resolve these.
3. **Ranking:** Select the "Best" baseline based on the highest fidelity to the visual evidence.

**Input Candidates:**
[Candidate 1]: {candidate_1_text}
[Candidate 2]: {candidate_2_text}
[Candidate 3]: {candidate_3_text}

**Strict Output Format:**
You must output your response in valid XML format only. No preamble, no markdown formatting outside the XML, and no conversational filler.

**XML Output Schema:**
<voting_result>
    <analysis>
        <candidate_1_critique>Briefly note strengths/hallucinations for C1.</candidate_1_critique>
        <candidate_2_critique>Briefly note strengths/hallucinations for C2.</candidate_2_critique>
        <candidate_3_critique>Briefly note strengths/hallucinations for C3.</candidate_3_critique>
    </analysis>
    <best_candidate_id>Candidate ID (1, 2, or 3)</best_candidate_id>
    <rationale>A concise explanation of why this candidate won, specifically citing why it outperformed the others in terms of accuracy or detail.</rationale>
</voting_result>
\end{lstlisting}
\end{tcolorbox}

\subsection{Evaluation Prompt}
We utilize \texttt{GPT-OSS-120B}~\citep{openai2025gptoss} to evaluate models' generated captions using the following prompts.

\begin{tcolorbox}[title={\textbf{Prompt for model evaluation}}, colback=teal!5!white, colframe=teal!75!black, boxrule=0.8pt, arc=2pt, breakable]
\begin{lstlisting}[basicstyle=\ttfamily\footnotesize, breaklines=true]
You are an expert Rubric Evaluator for Vision-Language Models.

### Task
Your task is to verify whether a model's generated **Caption** satisfies a specific set of **Rubrics** (Evaluation Criteria).
You will receive three inputs:
1.  **Model Caption:** The text description generated by the model.
2.  **Group A (Critical Rubrics):** A list of fundamental perception criteria. These are "bottom-line" facts.
3.  **Group B (Granular Rubrics):** A list of fine-grained or high-frequency error checks.

### Judgment Logic
For each rubric in both groups, determine if the **Model Caption** complies with the requirement.
*   **True (Pass):** The caption explicitly meets the criteria or implies it without ambiguity.
*   **False (Fail):** The caption contradicts the criteria, fails to mention a required element, or triggers a negative constraint.

**Handling Different Rubric Types:**
1.  **Positive Constraints** (e.g., "Must identify the car as red"):
    *   Pass: "A red car is parked..."
    *   Fail: "A blue car..." (Contradiction) OR "A car is parked..." (Missing specific detail).
2.  **Negative Constraints** (e.g., "Must NOT mention a dog"):
    *   Pass: "A cat sits on the mat." (No dog mentioned).
    *   Fail: "A dog and a cat..." (Hallucination detected).

### Crucial Requirement
You must evaluate **Group A** and **Group B** independently and return the results in separate lists. The order of boolean results in the output must strictly match the order of the input rubrics.

### Output Format
Return the result strictly in valid XML format. Do not use Markdown code blocks.
<Assessment>
  <GroupA>
    <Result>true</Result>
    <Result>false</Result>
    <!-- Add more Result tags matching the number of rubrics in Group A -->
  </GroupA>
  <GroupB>
    <Result>true</Result>
    <Result>true</Result>
    <!-- Add more Result tags matching the number of rubrics in Group B -->
  </GroupB>
</Assessment>

Please evaluate the following caption against the provided rubric groups.

[Model Caption]
{caption}

[Group A: Critical Rubrics]
{group_a_rubrics}

[Group B: Granular Rubrics]
{group_b_rubrics}
\end{lstlisting}
\end{tcolorbox}

\section{Human Annotation Feedback}
\label{sec:appendix_annotation}

To ensure the high quality of the benchmark, we involved human annotators in the loop. Given the extreme complexity of the images and the exceptional length of the golden captions (averaging 770.42 words), we employed the ``Model-Ensemble-Vote-then-Human-Refine'' pipeline. We utilized state-of-the-art multimodal models (specifically Gemini-3-Pro, GPT-5.2, and Seed-1.8) to generate initial drafts via a voting mechanism, followed by meticulous human verification.


Annotators reported that the AI-generated drafts were surprisingly sophisticated, significantly reducing the need for structural rewriting. However, the process introduced specific challenges regarding vigilance and fine-grained verification.

\paragraph{Hard Cases and Visual Nuances.}
The primary difficulty lay in \textbf{fine-grained visual semantic alignment}, particularly in regions with blurred edges, complex lighting, or severe occlusion. Annotators identified three recurrent types of ``hard cases'':
\begin{itemize}
    \item \textbf{Material and Boundary Misinterpretation:} Models occasionally merged ephemeral visual features with solid objects. A cited example involved a racing car where the model incorrectly described the ``dust kicked up by the wheels'' as a physical extension of the car's bodywork.
    \item \textbf{Precise Spatial Reasoning:} Subtle prepositional errors were common. For instance, a model described a pig as standing ``outside the pen,'' whereas a closer inspection revealed it was actually standing ``at the doorway'' (threshold ambiguity).
    \item \textbf{Hallucination in Low-Visibility Areas:} In shadowed or blurry regions, models tended to hallucinate specific, irrelevant objects to complete the scene.
\end{itemize}

\paragraph{Annotation Policy: Determinism over Ambiguity.}
Our annotators adhered to a strict standard of \textbf{determinism}. Unlike models that might produce vague descriptions for unclear regions (e.g., ``a blurry object''), humans preferred to \textbf{delete} hallucinations entirely rather than retaining ambiguous text. If an object was recognizable (e.g., via tool-assisted zooming), it was described explicitly; otherwise, it was removed to ensure the caption contained only grounded, high-confidence information.

\paragraph{Diversity of Caption Styles.}
Interestingly, annotators noted that the golden captions naturally exhibited distinct stylistic modalities, reflecting the versatile capabilities of the underlying models. The captions generally fell into two categories:
\begin{itemize}
    \item \textbf{Literary Narrative:} Highly fluent, prose-style descriptions that focus on immersion and flow. These captions tend to be exceptionally long and use varied sentence structures to weave visual details into a cohesive story.
    \item \textbf{Structured Representation:} Captions that utilize \textit{Markdown} formatting (e.g., bolding key terms, using bullet points for distinct regions) to present information in a highly organized, hierarchical manner.
\end{itemize}
We preserved this stylistic diversity in the final benchmark to evaluate models on both narrative generation and structured information extraction.

\section{Additional Experimental Results}
\label{sec:appendix_results}

Table~\ref{tab:results_mr_ew} presents the comprehensive evaluation results across all models.

\begin{table*}[h]
\centering
\caption{Main evaluation results on PerceptionRubrics. Models are categorized into Open-Source and Proprietary groups and sorted by Overall Score in \textbf{ascending order}. All values are reported in percentage (\%).
\textbf{M-R Item}: Must-Right Item Accuracy;
\textbf{E-W Item}: Easy-Wrong Item Accuracy;
\textbf{Gate Pass}: The sample-level pass rate where all Must-Right items are correct (Must-Right All True);
\textbf{E-W Avg}: The sample-level mean of per-case Easy-Wrong accuracy.}
\label{tab:results_mr_ew}
\resizebox{0.8\textwidth}{!}{
\begin{tabular}{l|c|cc|cc}
\toprule
\textbf{Model} & \textbf{Overall} & \textbf{M-R Item} & \textbf{E-W Item} & \textbf{Gate Pass} & \textbf{E-W Avg} \\
\midrule
\multicolumn{6}{c}{\textit{\textbf{Open-Source Models}}} \\
\midrule
Qwen2.5-VL-7B               &  8.37 & 64.99 & 30.69 & 26.20 & 30.52 \\
Qwen2.5-VL-32B              & 17.79 & 76.30 & 44.04 & 39.40 & 43.68 \\
Qwen3-VL-8B-Thinking        & 34.13 & 85.33 & 59.72 & 56.65 & 59.26 \\
Step3-VL-10B                & 35.97 & 85.63 & 59.05 & 58.96 & 58.95 \\
Qwen3-VL-235B-A22B-Thinking & 41.88 & 88.61 & 64.40 & 64.16 & 64.30 \\
MiniMax-M3                  & 44.82 & 89.21 & 66.40 & 65.70 & 65.94 \\
MiMo-V2.5                   & 45.65 & 88.49 & 65.77 & 66.76 & 65.57 \\
Step-3.7-Flash              & 48.62 & 90.39 & 70.04 & 68.21 & 69.80 \\
Kimi-K2.5                   & 50.78 & 91.43 & 71.31 & 71.10 & 70.94 \\
Kimi-K2.6                   & 51.77 & 91.45 & 71.04 & 72.16 & 70.92 \\
Qwen3.5-397B-A17B           & 61.61 & 93.64 & 78.01 & 78.90 & 77.59 \\
\midrule
\multicolumn{6}{c}{\textit{\textbf{Proprietary Models}}} \\
\midrule
GPT-4o-2024-05-13           & 12.59 & 70.01 & 36.00 & 32.27 & 35.67 \\
Seed-1.6             & 44.54 & 88.95 & 67.09 & 65.32 & 66.82 \\
GLM-5V-Turbo                & 48.18 & 90.84 & 68.79 & 69.17 & 68.60 \\
Seed-1.8                    & 54.34 & 91.69 & 72.12 & 73.60 & 71.91 \\
GPT-5.5                     & 55.23 & 93.21 & 69.34 & 78.23 & 69.47 \\
Gemini-3-Flash              & 59.83 & 93.25 & 75.44 & 78.23 & 75.01 \\
GPT-5.4                     & 60.81 & 93.60 & 74.88 & 79.58 & 74.73 \\
Seed-2.0-Pro                & 61.44 & 92.95 & 78.29 & 77.65 & 77.97 \\
Qwen3.5-Plus                & 61.61 & 93.86 & 77.86 & 79.09 & 77.40 \\
Qwen3.6-Plus                & 62.30 & 93.96 & 77.48 & 79.67 & 77.04 \\
Gemini-3-Pro                & 68.79 & 95.26 & 81.96 & 83.62 & 81.67 \\
Gemini-3.1-Pro              & 69.02 & 95.69 & 81.47 & 84.49 & 81.22 \\
Gemini-3.5-Flash            & 69.88 & 95.52 & 82.81 & 84.01 & 82.35 \\
Seed-2.0-Lite               & 70.07 & 95.59 & 84.69 & 82.85 & 84.23 \\
\bottomrule
\end{tabular}%
}
\end{table*}



\section{Qualitative Examples}
\label{sec:appendix_examples}
\begin{figure*}[t]
    \centering
    \includegraphics[width=0.9\textwidth]{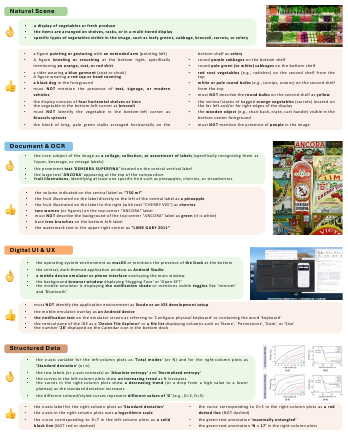}
    \caption{Qualitative examples of the fine-grained rubrics across four categories: Natural Scene, Document \& OCR, Digital UI \& UX, and Structured Data. Each example consists of an image and two tiers of rubrics: \textit{Must-Right} (top group) focusing on core facts, and \textit{Easy-Wrong} (bottom group) focusing on challenging details, negative constraints, and logical reasoning.}
    \label{fig:qualitative_cases_1}
\end{figure*}

\begin{figure*}[t]
    \centering
    \includegraphics[width=0.9\textwidth]{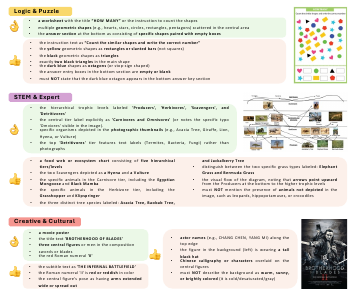}
    \caption{Qualitative examples of the fine-grained rubrics across three additional categories: Logic \& Puzzle, STEM \& Expert, and Creative \& Cultural. Each example consists of an image and two tiers of rubrics: \textit{Must-Right} (top group) focusing on core facts, and \textit{Easy-Wrong} (bottom group) focusing on challenging details, negative constraints, and logical reasoning.}
    \label{fig:qualitative_cases_2}
\end{figure*}

We provide concrete examples of the generated rubrics across diverse domains in Figure \ref{fig:qualitative_cases_1} and Figure \ref{fig:qualitative_cases_2}. 

As shown in the figures, our benchmark covers seven major categories, ranging from daily natural scenes to highly specialized STEM diagrams and logic puzzles. 
(a) For each image, we generate a comprehensive set of fine-grained rubrics. The items marked with the ``OK'' icon (Must-Right) represent core factual elements and primary subjects that are essential for a basic understanding of the scene. 
(b) The items marked with the ``Thumbs-up'' icon (Easy-Wrong) target more challenging details, including spatial relationships, fine-grained text recognition, negative constraints (e.g., ``must NOT mention...''), and complex logical reasoning. These rubrics are specifically designed to be ``Easy-Wrong'' for current large multi-modal models, effectively exposing hallucinations and subtle comprehension errors. For instance, in the ``Structured Data'' and ``STEM \& Expert'' cases, the rubrics require precise reading of axis scales, curve styles, and hierarchical biological relationships, which demand a high level of visual-logical alignment.

\end{document}